\documentclass[doc]{apa6}
\usepackage{apacite}
\usepackage{amssymb}
\usepackage{authblk}

\usepackage{times}
\usepackage{helvet}
\usepackage{courier}
\usepackage[cmex10]{amsmath}
\usepackage{amssymb,amsfonts, trsym}
\usepackage{graphicx}
\usepackage{tikz}
\usetikzlibrary{arrows}
\usetikzlibrary{automata,positioning}
\usepackage [english]{babel}
\usepackage [autostyle, english = american]{csquotes}
\MakeOuterQuote{"}
  
\usepackage{tikz}
\usepackage{filecontents}

\usepackage[utf8]{inputenc} 
\usepackage[T1]{fontenc}    
\usepackage{url}            
\usepackage{booktabs}       
\usepackage{amsfonts}       
\usepackage{nicefrac}       
\usepackage{microtype}      

\newcommand{\mat}[1]{\ensuremath{\mathbf{#1}}}

\newcommand{\Lk}{\ensuremath{\mathbf{L}^{\textrm{\scriptsize{-1}}}_{\textrm{k}}}}

\newcommand{\taustar}{\ensuremath{\overset{*}{\tau}}}

\newcommand{\ftilde}{\ensuremath{\tilde{f}}}

\DeclareRobustCommand{\tstr}{\ensuremath{\overset{*}{\tau}}}

\renewcommand{\ftilde}{\ensuremath{\tilde{f}}}
\newcommand{\ftildebold}{\ensuremath{\tilde{\mathbf{f}}}}

\newcommand\Editok[1]{}
\renewcommand{\Lk}{\ensuremath{\mathbf{L}^{\textrm{\scriptsize{-1}}}_{k}}}

\newcommand{\ii}{i}

\makeatletter \@ifundefined{@affil}{\def\@affil{~}{}}

\begin{document}

\title{Estimating scale-invariant future in continuous time}
\shorttitle{Scale-invariant future}

\author[1]{Zoran Tiganj}
\author[2]{Samuel J.~Gershman}
\author[3]{Per B.~Sederberg}
\author[1]{Marc W.~Howard}
\affil[1]{Center for Memory and Brain, Department of Psychological and Brain Sciences, Boston University}
\affil[2]{Department of Psychology and Center for Brain Science, Harvard University}
\affil[3]{Department of Psychology, University of Virginia}

\abstract{Natural learners must compute an estimate of future outcomes that follow from
a stimulus in continuous time. Widely used reinforcement learning algorithms discretize continuous time and  estimate either transition functions from one step to the next (model-based algorithms) or a scalar value of exponentially-discounted future reward using the Bellman equation (model-free algorithms). An important drawback of model-based algorithms is that computational cost grows linearly with the amount of time to be simulated. On the other hand, an important drawback of model-free algorithms is the need to select a time-scale required for exponential discounting. We present a computational mechanism,
developed based on work in psychology and neuroscience, for computing a
scale-invariant timeline of future outcomes. This mechanism efficiently
computes an estimate of inputs as a function of future time on a logarithmically-compressed scale, and can be used to generate a
scale-invariant power-law-discounted estimate of expected future reward.  
The representation of future time retains information about what
will happen when.
The entire timeline can be constructed in a single parallel operation which  generates concrete behavioral and neural predictions.
This computational mechanism  could be incorporated into future reinforcement learning algorithms.
}

\maketitle

\section{Introduction}
The ability to learn and operate in a continuously changing world with
complex temporal relationships is critical for survival. For example,
rats have to navigate around narrow holes and across wide fields, they have to
learn that some stimuli present imminent danger requiring quick action, while
others can serve as cues for events that will take place in a more distant
future.  Understanding the neural mechanisms that govern such behavioral
flexibility and building artificial agents that have such capacity poses a
significant challenge for neuroscience and artificial intelligence. 

In reinforcement learning (RL), an agent learns how to optimize its actions from
interacting with the environment. The traditional approach to RL is to consider
each different configuration of the environment as a different state
\cite{SuttEtal98}. 
Temporal difference (TD) learning has been employed to learn the scalar value of
temporally discounted expected future reward for each  
  state.  This approach
has been tremendously useful and led to  numerous practical applications (see
e.g., \citeNP{MnihEtal15}).  

In this paper we introduce a method for computing an estimate of future events along a logarithmically compressed timeline---an estimate of what will happen when in the future. This method addresses two major limitations of mainstream RL algorithms. First, because TD learning attempts to estimate an
integral over a function of future time, it discards detailed information about the time at which future
events are expected to take place. Of course, human decision-makers can reason about the time at
which future events will occur, leading many authors to augment the fast value computation supplied
by TD learning with a model-based system (see \citeNP{DawDaya14} for a review). The model-based
system is typically assumed to be slow; for some standard algorithms, the time taken to predict an
outcome $n$ steps in the future requires $n$ matrix operations. Second, because the goal of TD learning is to estimate the exponentially-discounted expected cumulative future reward, the method necessarily
introduces a characteristic time scale (Figure~\ref{fig:exp_vs_pl}a). If the delay associated with the to-be-learned
relationship is small compared to this scale, the behavior of the model will be dramatically different
than if it is large compared to this scale.\footnote{Similar
arguments can be made when eligibility traces are considered.}  In this paper, we present an alternative
method for predicting future outcomes in continuous time that addresses these limitations.

\subsection{Fixing a time scale limits flexibility}
Consider the task of designing an agent that will be deployed in a
realistic environment without additional intervention from the designer.
Successful performance on many tasks requires the ability to learn
across a range of time scales.  To make this example more concrete, consider designing
an agent that will be deployed on the streets of Boston to learn to complete
the everyday tasks of a post-doc. In order to get from Boston
University to Harvard, the agent must learn that
switching onto the red line leads to Harvard Square about 20~minutes
in the future.  At Dunkin Donuts, the agent must learn that
paying money leads to a cup of coffee in about a minute. Grasping the cup
and sipping the coffee predicts the taste of coffee immediately, but predicts the
stimulating effect of caffeine several minutes in the future.  In designing an
agent to learn all of these tasks in an unknown environment,
the designer will not necessarily  know  what temporal scales are important.
We thus desire that the learning algorithm be scale-invariant.

\begin{figure*}
\centering
\begin{tabular}{lc}
\textbf{a} & \\
&		\includegraphics[width=0.5\textwidth]{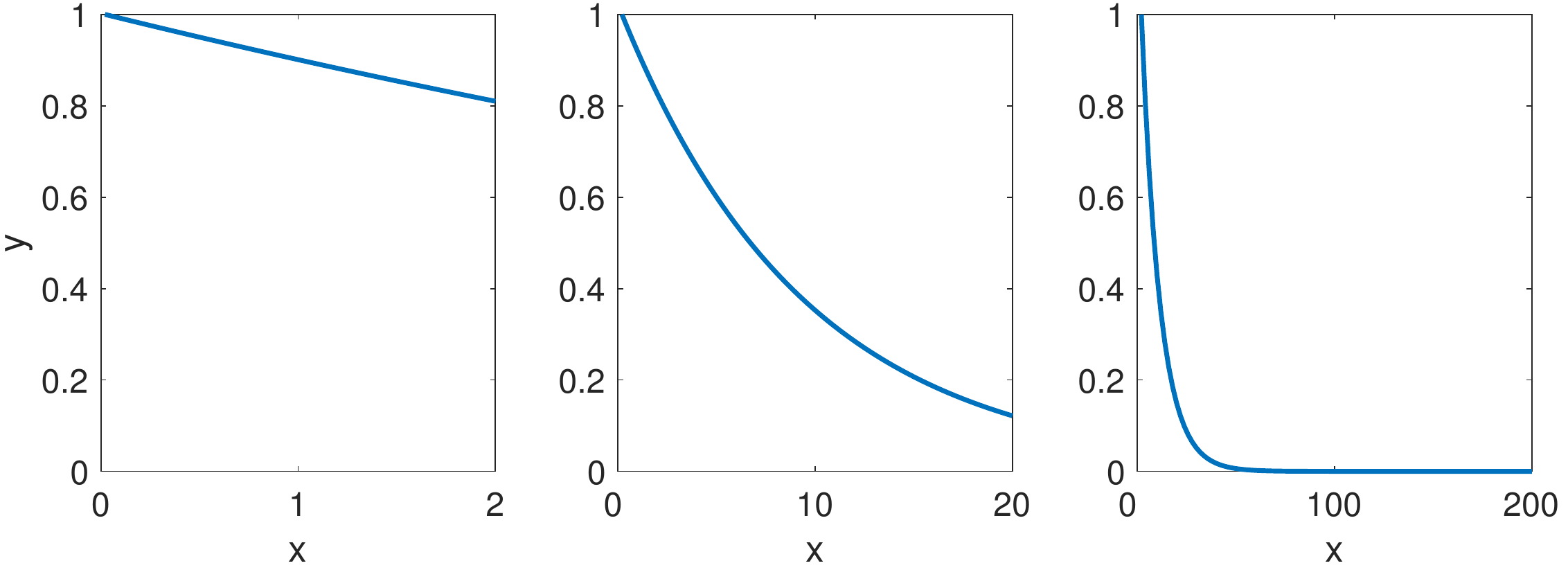}\\
\textbf{b} &\\
&		\includegraphics[width=0.5\textwidth]{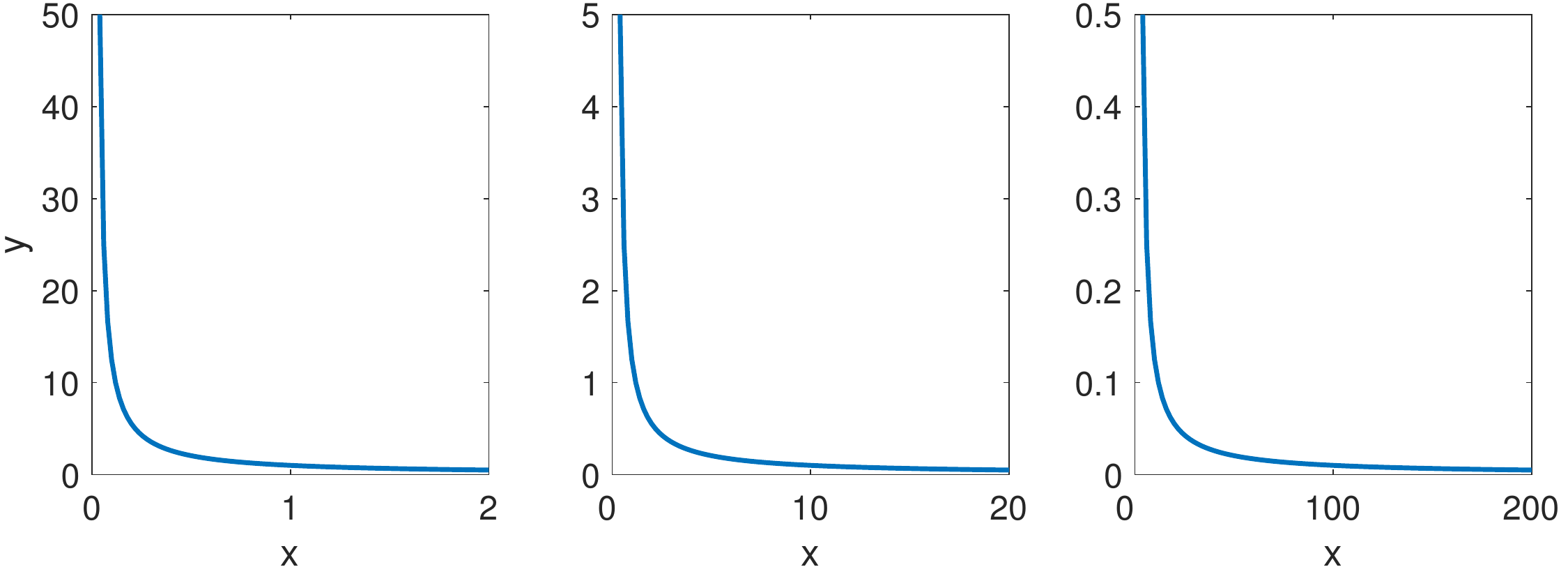}	
 \end{tabular}\\ \caption{Exponential discounting introduces a scale;
		 power-law discounting is scale-free. { \textbf{a.}
		 An exponential function has qualitatively different properties at
		 different scales.  The function $y = \gamma^x$ is shown at three
		 different scales for $\gamma= 0.9$.
		If $x$ is on the order of the time constant ($-\log \gamma$)   we
		obtain the familiar exponential function with a clearly defined
		gradient for different values of $x$  (middle).  If $x$ is small with
		respect to the time constant, we find a linear function with a shallow
		slope (left).
		If  $x$ is large relative to the time constant (right), the function
		approximates a delta function with a peak around zero. 
		 \textbf{b.} Power-law
		 discounting ($y = x^{-1}$).  For all ranges of $x$ values the power
		 law gives the same relative gradient of values.
		 \label{fig:exp_vs_pl}
}  }
\end{figure*}

Algorithms based on the Bellman equation, which includes TD learning,
estimate an exponentially-discounted expected future return (value) by harnessing the recursive structure of the value function $V(t)$:
\begin{align}
V(t) = \mathbb{E}[r(t) + \gamma V(t+1)],
\end{align}
where $r_t$ denotes the reward at time $t$, the expectation represents an average over future events, and the exponential discount factor $\gamma$ fixes a characteristic time
scale.\footnote{
More precisely, the inverse of the time constant goes like $-\log \gamma$.
}
The scaling results in very different policies at different temporal scales
 (Figure~\ref{fig:exp_vs_pl}a).  Consider a world in which two rewards $A$ and
$B$ follow a cue.  The delay from the cue to $B$ is twice the delay to $A$.
Suppose that we do not know the units of time in the world and pick 
$\gamma = 0.9$.
If the units of the world are such that the delay to $A$ is 1  and the delay
to $B$ is 2, then the agent would prefer $B$ if the value of  $A$ was \$$1$
and $B$ was
\$$1.2$. However, if the units of the world are very different such that 
the delay to $A$ was 100 and the delay to $B$ was 200, then even if the
reward at $B$ was \$$30,000$, the agent would still prefer $A$.  This
example makes clear that the success of a model that makes use of exponential
discounting depends critically on aligning the choice of $\gamma$ to the
relevant scale of the world.  In addition, animal literature suggests that
hyperbolic discounting explains the data better than exponential discounting
(see e.g. \citeNP{GreeMyer96}), for instance regarding preference reversal
\cite{GreeMyer04,Hayd16}.

\subsection{Representing the future with a scalar obscures temporal information}
One could implement scale-invariant power-law discounting\footnote{If $f(t) =
t^a$, then rescaling the time axis preserves the relative values at all time
points, $f(\alpha t) = C f(t)$.  } by choosing an appropriate spectrum of
exponential discount rates \cite{KurtEtal09,Sutt95}.	
However it is computed, a discounted value discards potentially important
information about when an anticipated event will occur. 
For instance, consider the decision facing an agent about whether to buy a
cup of very hot coffee.  Drinking the coffee immediately would burn one's
mouth.  However, drinking the
coffee after waiting a few minutes for it to cool down will result in a
delicious and stimulating beverage.  Is the value
of the coffee negative (burned mouth) or positive (delicious beverage) or some
weighted sum of the two?  
One way to answer the question is to state that the value of the coffee is a
function over future time that is initially negative and then later positive.
If the only information about this function  that can be brought to bear in
deciding whether to purchase the coffee is a single scalar value, then the
decision-maker may choose an inappropriate action, either 
purchasing the coffee when she does not have time to wait for it to cool or 
missing the opportunity to enjoy a delicious beverage in the 
  near future.


One could tackle this problem using \textit{model-free} RL approaches by expanding the state space to include relevant variables, such as the elapsed time that the cup has been held and the steam coming from the cup. However, when the elapsed time is one of the variables that the agent needs to keep track of, this approach becomes computationally very expensive. This is because the number of states rapidly increases. If time is discretized into $m$ bins and we need to keep track of $n$ stimuli, then the number of states is $m^n$. This is especially costly when time is discretized in equal-sized bins as in complete serial compound representation. Using microstates characterized with a set of compressed temporal basis functions as in \cite{LudvEtal08,LudvEtal12} reduces the number of states to some degree, but this type of representation does not provide a future timeline.

Classical \emph{model-based} RL enables decisions that take into account the
time at which future events will take place. However the computational cost of
traditional model-based solutions grows linearly with the horizon over which
one needs to estimate the future. In this paper, we present a method that
constructs a function over future time for each stimulus (state).  This representation
of the future is logarithmically compressed and the estimate of the future at
many different points in time can be computed in parallel.  One could compute
an integral over this representation to maintain a cached value with power-law
discounting.  But because the entire function is available, an agent can also
incorporate the time at which rewards will become available into its
decision making.  

\subsection{Scale-invariant temporal representations in the brain}


The basic computational strategy we pursue is to 1) compute a
scale-invariant representation of the temporal history leading up to the
present and 2) at each moment associate the history  with the stimulus
observed in the present.  Step~1 assumes the existence of a scale-invariant
compressed representation of temporal history.  Step~2 assumes the existence
of an associative mechanism.  There is ample neural evidence for both of these
assumptions.  A large literature from the cellular neuroscience literature
provides evidence for an associative mechanism implementing Hebbian plasticity
at synapses \cite{BlisColl93,LismEtal02}, which would be required for Step~2.

There is also a growing body of evidence consistent with assumptions necessary
for Step~1. Experiments from several species suggest that the brain maintains a
compressed representation of time in multiple brain regions. ``Time cells''
fire during a circumscribed period of time within a delay interval
\cite{PastEtal08,MacDEtal11}; a reliable sequence of time cells tile  the
delay on each trial (Figure~\ref{fig:heatmaps}a).  Because the
sequence is reliable, time cells can be used to reconstruct how long in the
past the delay began.  In many experiments, these sequences also carry
information about what stimulus initiated the delay interval
\cite{PastEtal08,MacDEtal13,TigaEtal18a,TeraEtal17}.
Because there are fewer cells that fire later in the
sequence and those that fire later in the sequence fire for a longer duration
\cite{HowaEtal15,SalzEtal16}, the ability to reconstruct time decreases as
the start of the interval recedes into the past.  Time cells have been
observed in several brain regions, including hippocampus
\cite{MacDEtal11,SalzEtal16}, prefrontal cortex
\cite{TigaEtal16,BolkEtal17,TigaEtal18a} and striatum
\cite{MellEtal15,AkhlEtal16}, in several species
\cite{MauEtal18,AdleEtal12,TigaEtal18a} and in a wide variety of behavioral
tasks.  

Taken together, these data indicate that at each moment the brain maintains a
temporal record of what happened when leading up to the present.  The decrease
in accuracy for events further in the past suggests that this temporal record
is compressed.  
As such, this neural data aligns with longstanding predictions from
cognitive models
\cite{BrowEtal07,BalsGall09,HowaEtal15}.  
These models further predict that the form of compression should be
logarithmic.
Behavioral models built from a logarithmically-compressed representation 
readily account for scale-invariant behavior \cite{HowaEtal15}.\footnote{For
		much the same reason that, on a logarithmic scale, the difference
		between 1~and~2 is the same as the difference between 100 and 200,
		models built from a logarithmically-compressed temporal representation
will be scale-invariant.}


\subsection{Overview of this paper}
In this paper we use a logarithmically-compressed record of the
past---a set of appropriate time cells---to construct a scale-invariant
estimate of the time of \emph{future} events.  A logarithmically-compressed
record of the past can be efficiently computed using a
method we will describe in detail below \cite{ShanHowa12,ShanHowa13}.
At each moment, this representation of the past 
is associated to the present.  Neurally, this association requires nothing
more elaborate than Hebbian plasticity, which can be implemented \emph{via}
long-term potentiation \cite{BlisColl93}.
The past-to-present association can also be
understood as a present-to-future association.  As
such, multiplying this association with the present stimulus vector enables us to
identify the sequence of stimuli that will \emph{follow} the probe stimulus at
different points in the future.   Section~\ref{sec:method}  describes this
method more precisely.

This method yields an estimate of the future that has very different
properties than traditional approaches used in RL.  The properties of this
representation are described with illustrative examples in
Section~\ref{sec:examples}.
Because the representation of the past is logarithmically compressed, so too
is the estimate of the future that it produces.  
A cached scalar value can be computed from this timeline, yielding
(scale-invariant) power-law discounting by summing over the predicted future. 
Notably, the compressed timeline representation also provides a function over
simulated time.  The future timeline can be computed in a single parallel
operation and sums over potential outcomes. Section~\ref{sec:predictions}
describes neural and behavioral predictions of the model, reviewing recent
empirical results that are consistent with the proposed hypothesis that the
brain constructs a  logarithmically compressed future. 

\section{Constructing a logarithmically-compressed timeline of the future}
\label{sec:method}

This approach requires two key components,  a logarithmically-compressed memory
representation and an associative memory between the compressed representation
and the present stimulus.  Subsection~\ref{sec:compressed} describes a method
for constructing a logarithmically-compressed memory representation following
\citeA{ShanHowa13}.  Subsection~\ref{sec:associative} describes the
associative memory.  Subsection~\ref{sec:predict} describes the future
timeline that results from probing the associative memory with a stimulus
representation.

\subsection{Previous work: Constructing  a compressed memory representation of the past}
\label{sec:compressed}

\begin{figure*}
\centering
\includegraphics[width=0.13\textwidth]{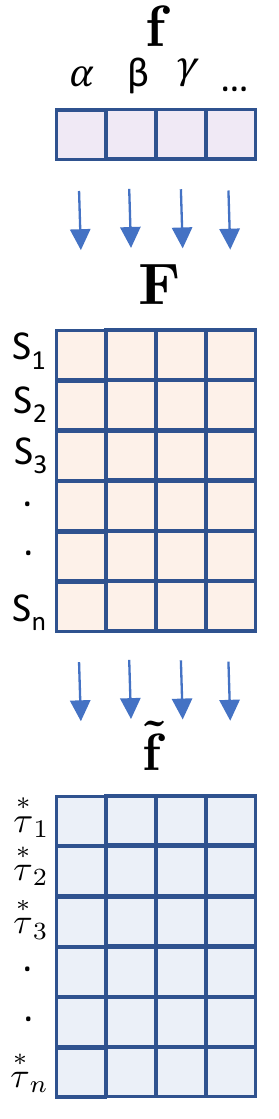}
 \caption{\small Constructing a memory representation of the recent past. The
		 schematic shows the two-layer network for constructing a memory
		 representation by implementing an approximation of the Laplace and
		 the inverse Laplace transform. The input $\mathbf{f}$ is a vector
		 over states $\alpha$, $\beta$, $\gamma$... This provides input to a
		 two-layer network where each layer  is a 2-D array (sheet) of
		 neurons. Neurons in the first layer $\mathbf{F}$ are leaky
		 integrators indexed by the state they encode, $\alpha$, $\beta$,
		 $\gamma$... and their rate constant
		 $s$.  We refer to the activation of a particular entry 
		 as $F_{s,\alpha}$.  Neurons in the second layer $\ftildebold$
		 activate sequentially following the input stimulus. They are indexed by
		 the state that provides their input and the time by which the peak of
		 their activation follows the input stimulus $\taustar$.   The activation
		 of a particular unit is referred to as $\ftilde_{\taustar,\alpha}$.
			 \label{fig:indices}}
 \end{figure*}

Consider a case in which the network is presented with a vector-valued input
that changes over time $\mathbf{f}(t)$.  This input reflects the presence or
absence of a set of discrete stimuli (states) that we denote as $I={\alpha, \ \beta, \ \gamma \ldots}$.  For simplicity, let us assume that the input uses
a localist (one-hot) representation; 
if stimulus $\alpha$ is present at time $t$ we write $f_{\alpha}(t)=1$.  Now,
the goal of this method is to construct an
estimate of the past  leading up to the present. 
We refer to this memory representation as $\ftildebold$. 
A temporal record of the past requires two types of
information.  In order to estimate $\mathbf{f}(t' < t)$ we need to
maintain both what and when information.
Thus, we index each of the neurons in $\ftildebold$ by two
indices, $\ftilde_{\taustar, \ii}$ (Figure~\ref{fig:indices}). 
The second index $\ii \in I$ corresponds to the what information.  The other index, $\taustar$ refers to the time in the past
that this neuron is attempting to represent.
That is, the network includes a set of values of $\taustar \in 
\{\taustar_1, \ \taustar_2, \ \taustar_3$ \ldots$\}$.  Because the value of $\taustar$ for the $i$th row of the
network, $\taustar_i$ has physical meaning, we refer to the neurons in
$\ftildebold$ by their value of $\taustar$ rather than their row number.  
Each entry $\ftilde_{\taustar,\alpha} (t)$ approximates
$f_\alpha(t + \taustar)$.  Here the values of $\taustar$ are negative as they
refer to a temporal distance in the past relative to the present. 

\begin{figure*}
	\begin{center}
			\begin{tabular}{lc}
			\textbf{a} & \\
		&		\includegraphics[width=1\textwidth]{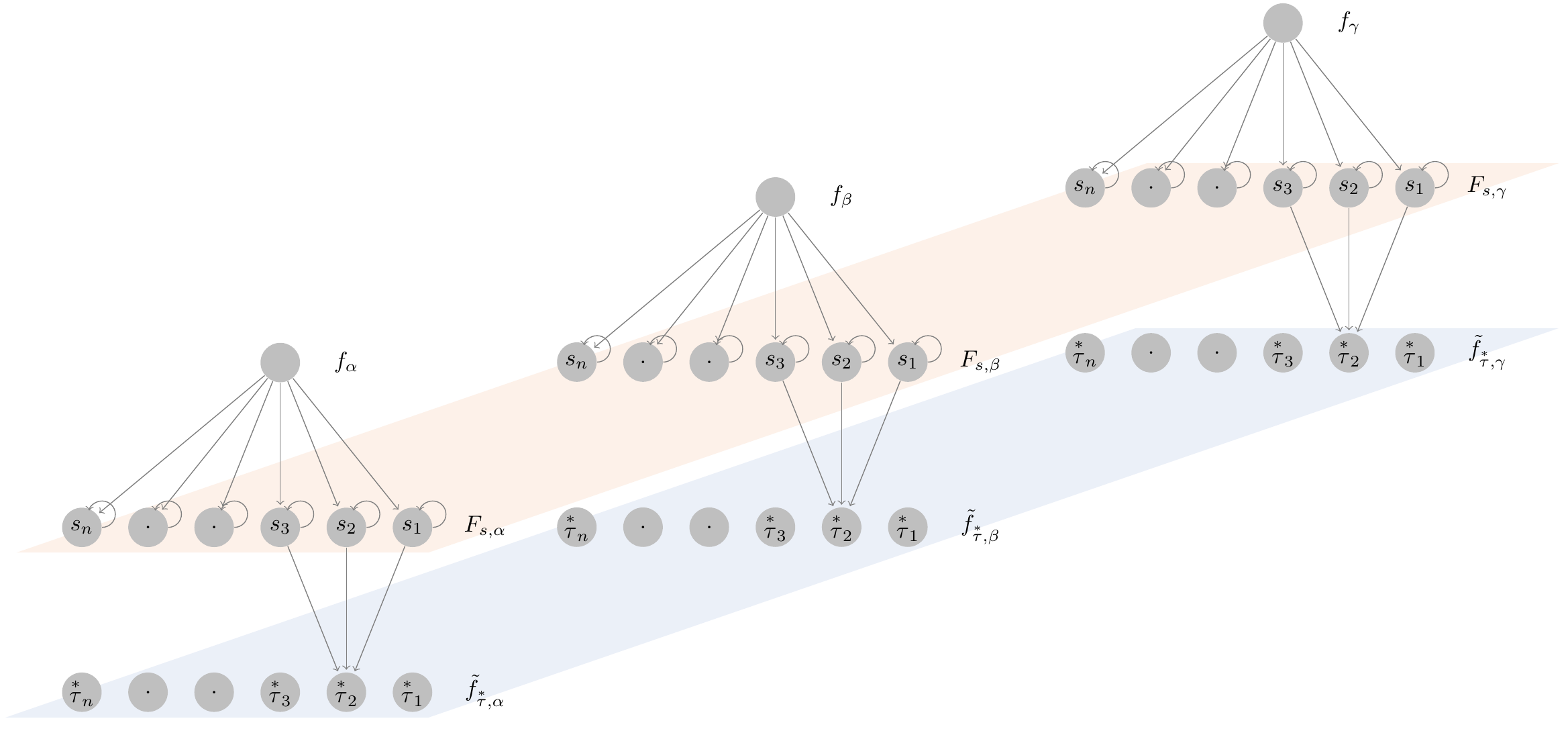}
		 \end{tabular}\\
			\begin{tabular}{lll}
				\textbf{b} &
				\textbf{c} \\
				\includegraphics[width=0.4\linewidth]{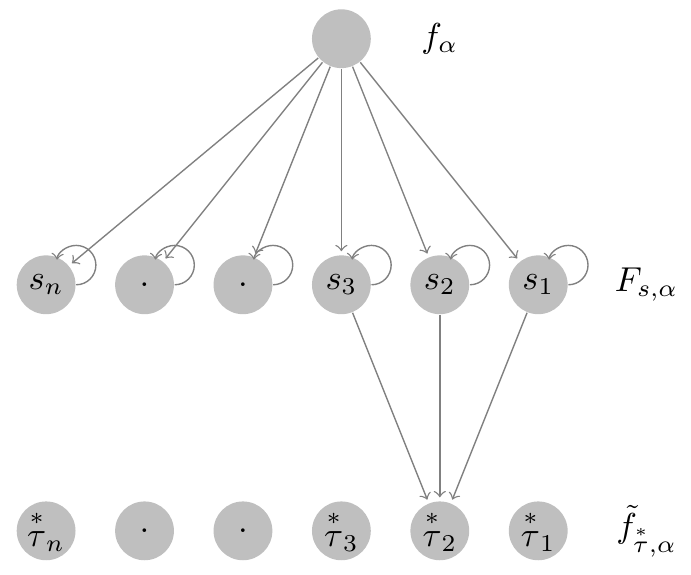}
				&
				\includegraphics[width=0.48\linewidth]{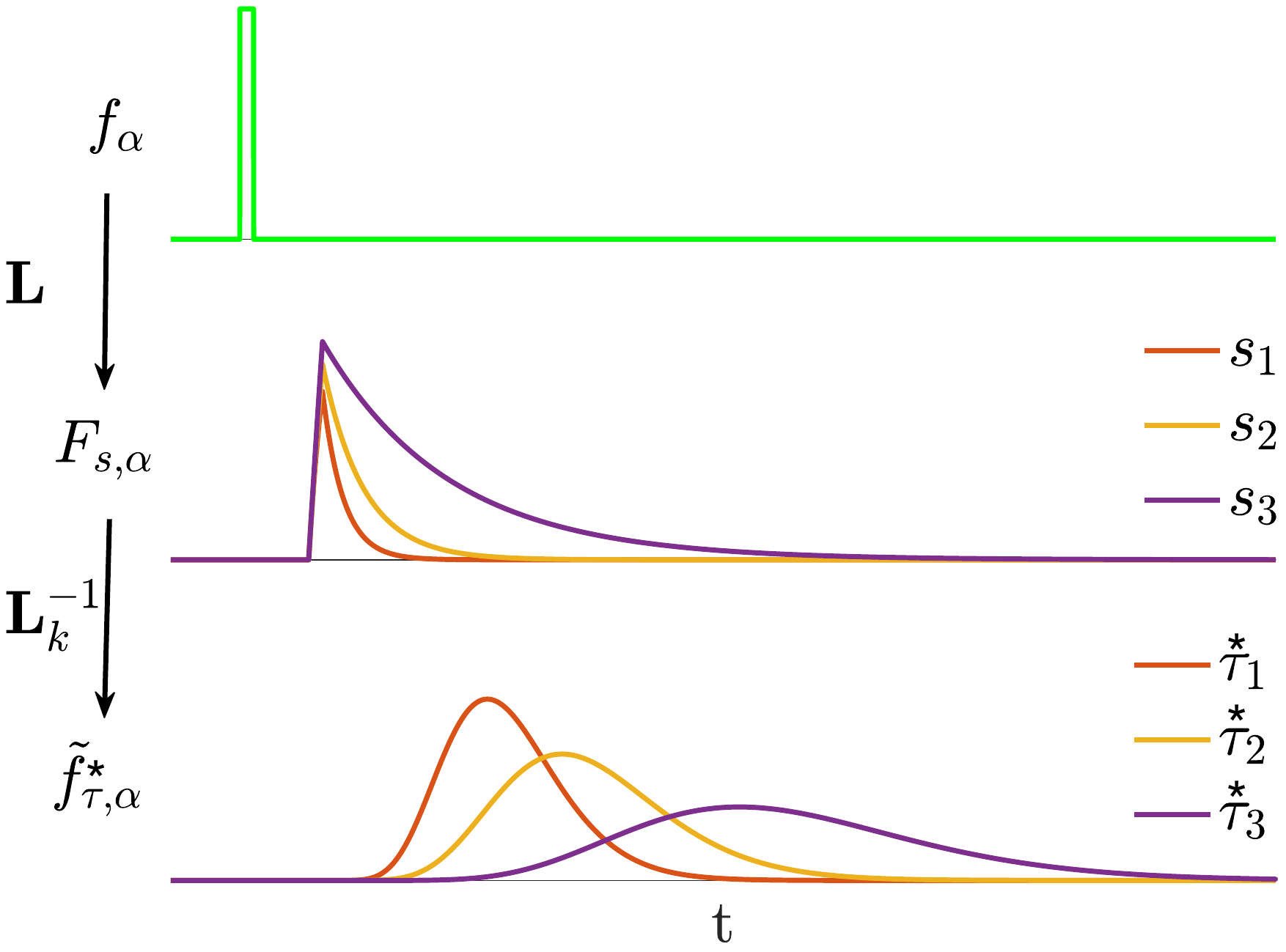}\\
			\end{tabular}
	\end{center}
	\caption{\small Structure and dynamics of the memory representation.
			\textbf{a.} The two-layer network is organized such that each
			input state has its own set of units in $\mathbf{F}$ and
			$\ftildebold$ layers (when constructing the memory representation,
			there is no crosstalk between the neurons that correspond to
			different states). \textbf{b.} The input stimulus $f_{\alpha}(t)$
			feeds into a layer of leaky integrators $F_{s,\alpha}$ that
			implement a discrete approximation of an integral transform.
			Each neuron in the first layer has a characteristic rate constant
			$s_i$.
 $F_{s,\alpha}$ projects onto
			$\tilde{f}_{\taustar,\alpha}$ through a set of weights defined with the
			operator $\Lk$ which implements an approximation of
			the inverse of the Laplace transform. Notice that the $\Lk$
			operator projects only to a local neighborhood ($k$ neurons).  Neurons in the second layer each have
			their characteristic peak time relative to the input onset
			$\taustar_i$. Analytic relationship between $\taustar$ and $s$ can be
			expressed as $\taustar = -k/s$. Thus choosing $\taustar$ and integer $k$
			fully specifies $s$, similarly, choosing $s$ and $k$ fully
			specifies $\taustar$. We chose $\taustar$ to be logarithmically spaced
			(in order to have a logarithmically compressed memory
			representation). 			
			\textbf{c.} A response of the network to a delta-function input. 
			Activity of only  three neurons in each layer is shown.  
			Neurons in $\tilde{f}_{\taustar,\alpha}$ activate sequentially
			following the stimulus presentation.  The width of the activation
			of each neuron scales with the peak time determined by the
			corresponding $\taustar$, making the memory scale-invariant.
			 \label{fig:F_and_f}}
	  \end{figure*}
	  
Following prior work \cite{ShanHowa12,ShanHowa13} we will construct the
representation of the past $\ftildebold$ by means of an intermediate
representation $\mathbf{F}$.  Each neuron in $\mathbf{F}$ aligns with a
corresponding neuron in $\ftildebold$ (Figure~\ref{fig:F_and_f}a).   The neurons in $\mathbf{F}$ are
indexed by the label of the stimulus in the world that activates
them ($\alpha, \ \beta, \ \gamma \ \ldots$)  and a scalar value $s$. 
The values of $s$ for each row of $\mathbf{F}$ align with the corresponding
values of  $\taustar$ in each row of  $\ftildebold$  (Figure~\ref{fig:F_and_f}b):
\begin{equation}
		F_{s,\alpha} \leftrightarrow \ftilde_{\taustar,\alpha}.
		\label{eq:Sheet}
\end{equation}
The mapping between  $s$ and $\taustar$ is such that  $s = -k/\taustar$ ,
where  $k$ is an integer with physical meaning that will be described below and $i \in 1, 2, 3 \ldots n$, where $n$ is the number of rows in $\mathbf{F}$ and $\ftildebold$.
As with $\taustar$, there are a finite set of
values of $s$, $s \in 
\{s_1, \ s_2, \ s_3$ \ldots$\}$. As with 
$\taustar$, there is a physical meaning to the $i$th value of $s$ so we refer
to neurons in
$\mathbf{F}$ by their value of $s_i$ rather than their index $i$.  
Values of $s$ are defined to be positive.  Following previous work
\cite{ShanHowa13,HowaShan18}, we choose the values of $\taustar$ and $s$ to be
evenly spaced on a logarithmic scale.\footnote{For instance, one can choose
$\taustar_i = \taustar_{\textnormal{min}}(1+c)^{i-1}$ for some
minimum value of $\taustar$, $\taustar_{\textnormal{min}}$ and a constant $c$
that controls the spacing.}

The dynamics of each unit in $\mathbf{F}$ obeys:
\begin{equation}
		\frac{dF_{s,\alpha}(t)}{dt} = -s F_{s,\alpha}(t) +
		{f}_\alpha(t),  \label{eq:Laplace}
\end{equation}
where the value of $s$ on the rhs refers to that particular neuron's value 
$s_i$.  Here we can see that $s$ describes each neuron's rate constant;
$1/s$ describes each neuron's time constant.
Taking the network state across all values of $s$, $\mathbf{F}(s)$ estimates the
Laplace transform of
$\mathbf{f}(t' < t)$.  
To see that $F_{s,\alpha}$ at time $t$ is the Laplace transform of
$f_{\alpha}(t' < t)$, solve Eq.~\ref{eq:Laplace}:
\begin{equation}
		F_{s,\alpha}(t) = 
			\int_{-\infty}^t e^{-s\left(t - 
		t' \right)}f_{\alpha}(t') dt'.
		\label{eq:solution}
\end{equation}

Knowing that $\mathbf{F}$ at time $t$ holds the Laplace transform of
$\mathbf{f}$ leading up to the present suggests a strategy to construct an
estimate of $\mathbf{f}$.  If we could invert the transform and write the
answer into another set of neurons $\ftildebold$, this would provide an
estimate of $\mathbf{f}$ as a function of time leading up to the present.
The Post approximation \cite{Post30} provides a recipe for approximating the
inverse transform that can be computed with a set of feedforward weights,
which we denote $\Lk$:
\begin{equation}
\ftilde_{\taustar,\alpha}(t) = \Lk F_{s,\alpha}(t).
		\label{eq:componentinverse}
\end{equation}
The integer $k$ determines the precision of the approximation. Denoting the
$k^{th}$ derivative with respect to $s$ as
$F^{(k)}_{s,\alpha}$ we can rewrite Eq.~\ref{eq:componentinverse} as:
\begin{equation}
	\tilde{f}_{\taustar,\alpha}(t) = 
		C_k s^{k+1}F^{(k)}_{s,\alpha}(t),
\end{equation}
where $C_k$ is a constant that depends only on $k$.


To get an intuition into the properties of $\ftilde$, we
present a delta function to $f_{\alpha}$ at time zero and examine the activity
of $F_{s,\alpha}(t)$ and 
$\ftilde_{\taustar,\alpha}(t)$.   We find immediately that $F_{s,\alpha}(t) =
e^{-st}$.  Moreover, the activity of the neurons in $\ftildebold$ obeys:
\begin{equation}
	\label{eq:imp_res}
	\tilde{f}_{\taustar, \alpha}(t) = 
		C_k \frac{1}{\taustar} \left(\frac{t}{\taustar}\right)^k
		e^{-k\frac{t}{\taustar}},
\end{equation}
where $C_k$ here is a different constant that depends only on $k$.
The activity of each node in $\ftilde_{\taustar, \alpha}$ is the product of an
increasing power term $\left(\frac{t}{\taustar}\right)^k$ and a decreasing 
exponential term $e^{-k\frac{t}{\taustar}}$. 
In the time following a delta function input, the firing of each neuron in
$\ftilde_{\taustar, \alpha}$ peaks at $\taustar$  (Figure~\ref{fig:F_and_f}c).
Thus, following a transient input of state $\alpha$, neurons in
$\ftilde_{\taustar,\alpha}$ activate sequentially. 

\begin{figure*}
\centering
\begin{tabular}{ll}
\textbf{a}  &
\textbf{b} \\
		\includegraphics[width=0.4\textwidth]{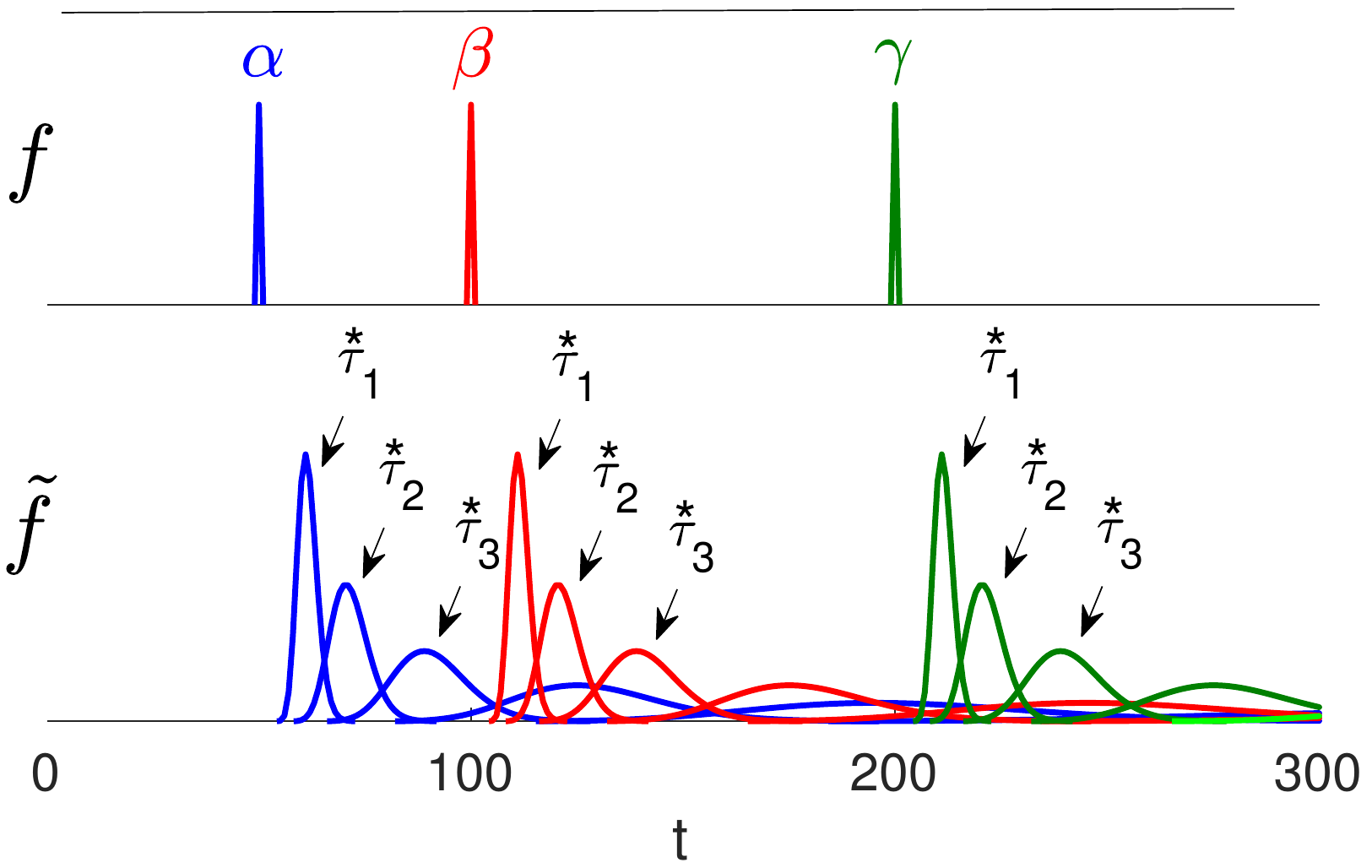}	
&		\includegraphics[width=0.35\textwidth]{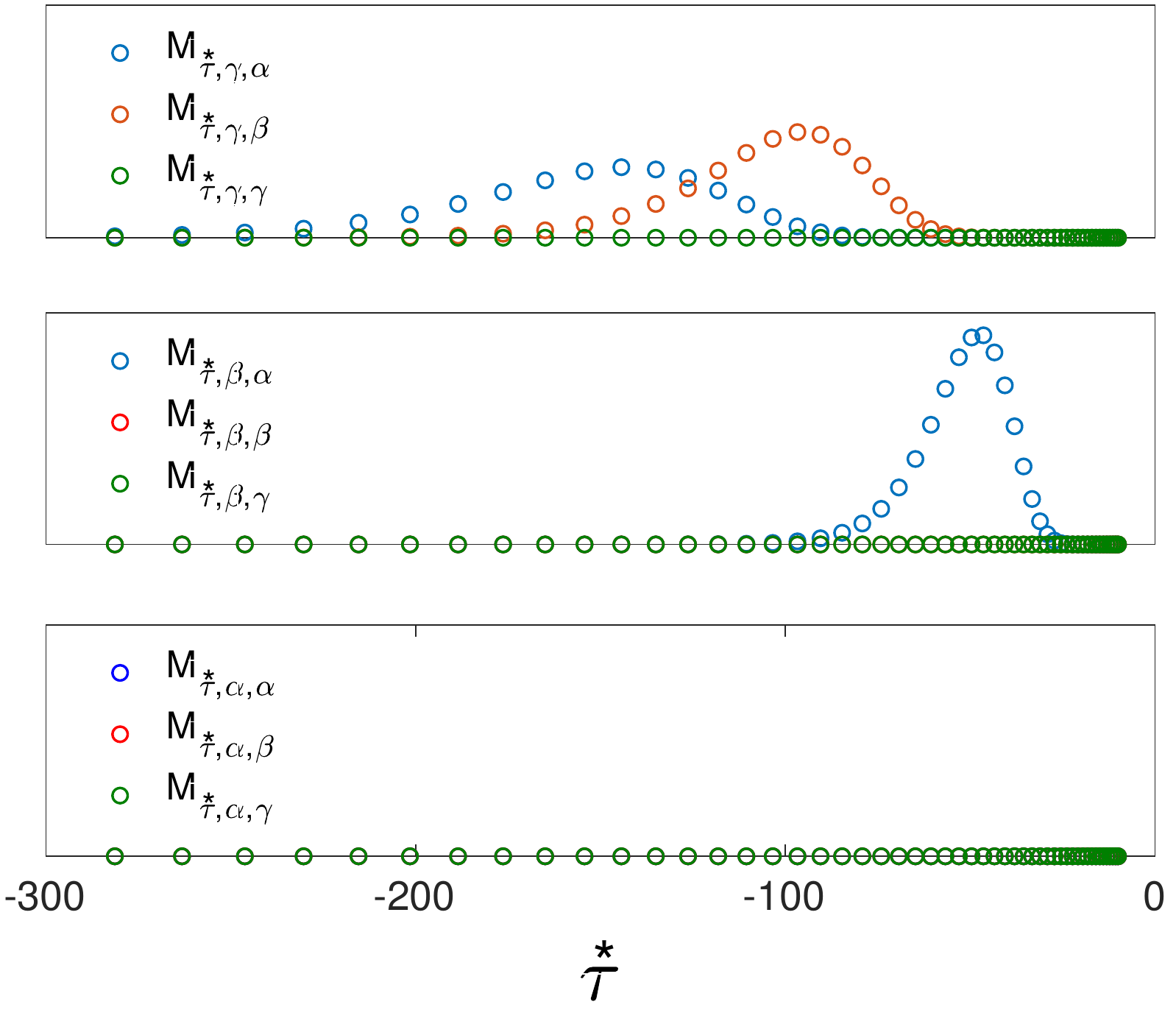}	
 \end{tabular}
 \caption{
 Constructing an associative memory by building connections between present
 inputs and memory of the recent past.  {\textbf{a.} Illustration of the
 compressed memory representation $\ftildebold$ as a function of time during
 presentation of the sequence $\alpha$, $\beta$, $\gamma$.  Stimuli
 presented at different times (top) induce sequential activation (bottom) in
 $\ftildebold$. Activation corresponding to different stimuli is shown with
 different colors. For clarity, only a handful of neurons are displayed.
 \textbf{b.} A graphical depiction of the state of the associative memory
 $\mathbf{M}$ after learning of the sequence $\alpha$, $\beta$, $\gamma$.
 Each of the three plots shows entries stored in $\mathbf{M}$ as a function of
 $\taustar$.  $\mathbf{M}_{\tstr,q,r}$ is the synaptic weight from $r$ to $q$ at a particular $\tstr$.
The different stimuli are shown in different colors in each plot.
			Top: Associations stored in $M_{\taustar,\gamma,\alpha}$,
			$M_{\taustar,\gamma,\beta}$ and $M_{\taustar,\gamma,\gamma}$ as  functions of log-spaced $\taustar$. Because $\gamma$
			was preceded by both {$\alpha$} and {$\beta$}, both
			$M_{\taustar,\gamma,\alpha}$ and $M_{\taustar,\gamma,\beta}$  have peaks.
			Because $\beta$ had been presented more recently when $\gamma$
			was presented, the curve for  $M_{\taustar,\gamma,\beta}$ has a peak
			closer to $\taustar=0$.  Because the representation of times in
			the more recent past is more accurate than times further in the
			past, the peak for $\beta$ is also more sharp as a function of
			$\taustar$. Middle: Associations stored in
			$M_{\taustar,\beta,\alpha}$, 
			$M_{\taustar,\beta,\beta}$ and $M_{\taustar,\beta,\gamma}$. Because $\beta$ was preceded by $\alpha$ at a short
			lag, $M_{\taustar,\beta,\alpha}$  differs from zero at low values
			of $\taustar$. Since $\beta$ was not  preceded by itself or by
			$\gamma$, blue and green traces are zero. Bottom: Associations
			stored in $M_{\taustar,\alpha,\alpha}$,
			$M_{\taustar,\alpha,\beta}$ and
			$M_{\taustar,\alpha,\gamma}$. Since $\alpha$ was not
			preceded by any stimulus, corresponding entries in $\mathbf{M}$
			are zero for all values of $\taustar$.
\label{fig:f_and_M}}}
\end{figure*}

\begin{figure*}
\begin{tabular}{ll}
		\textbf{a} &\\
	& 		\includegraphics[height=0.16\textheight]{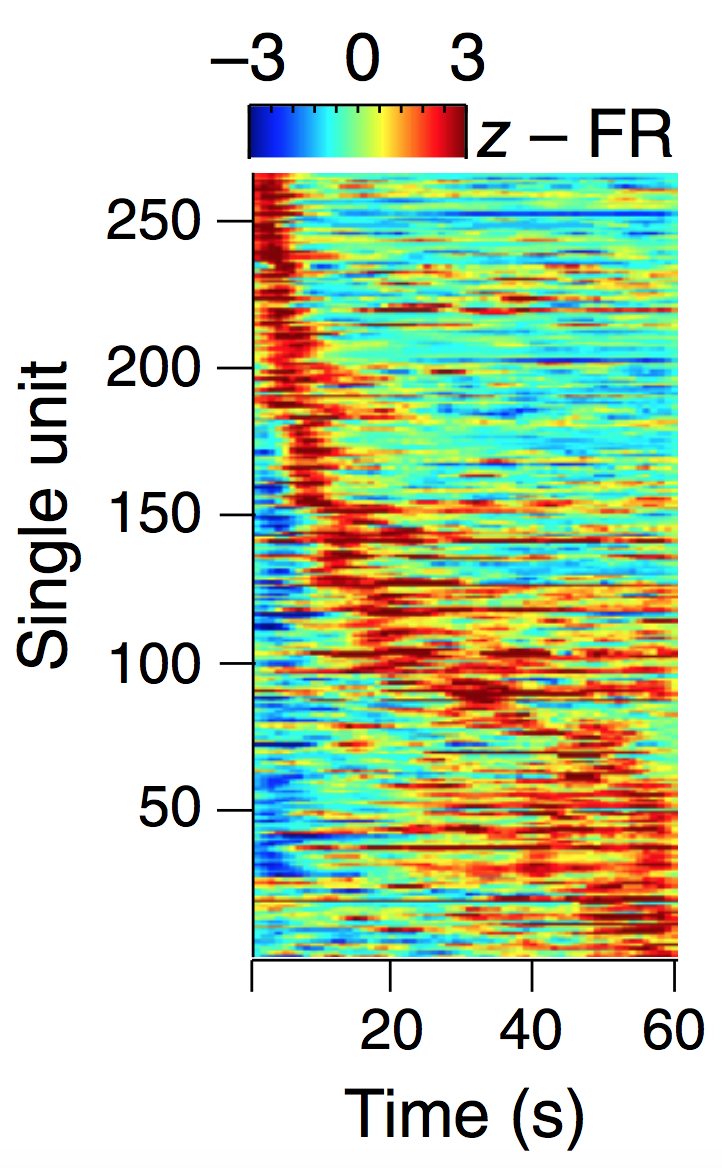}
	\includegraphics[height=0.16\textheight]{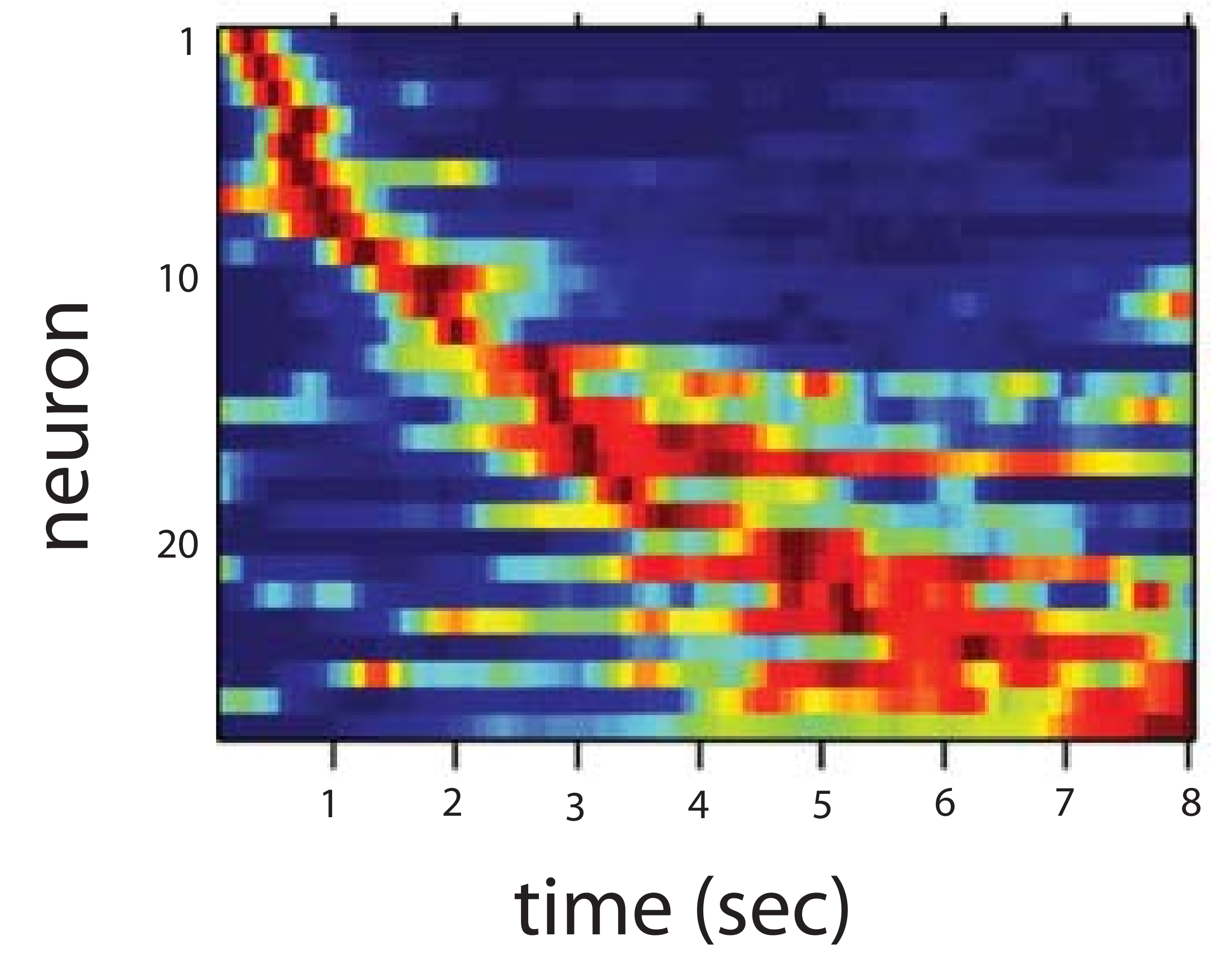}
	\includegraphics[height=0.16\textheight]{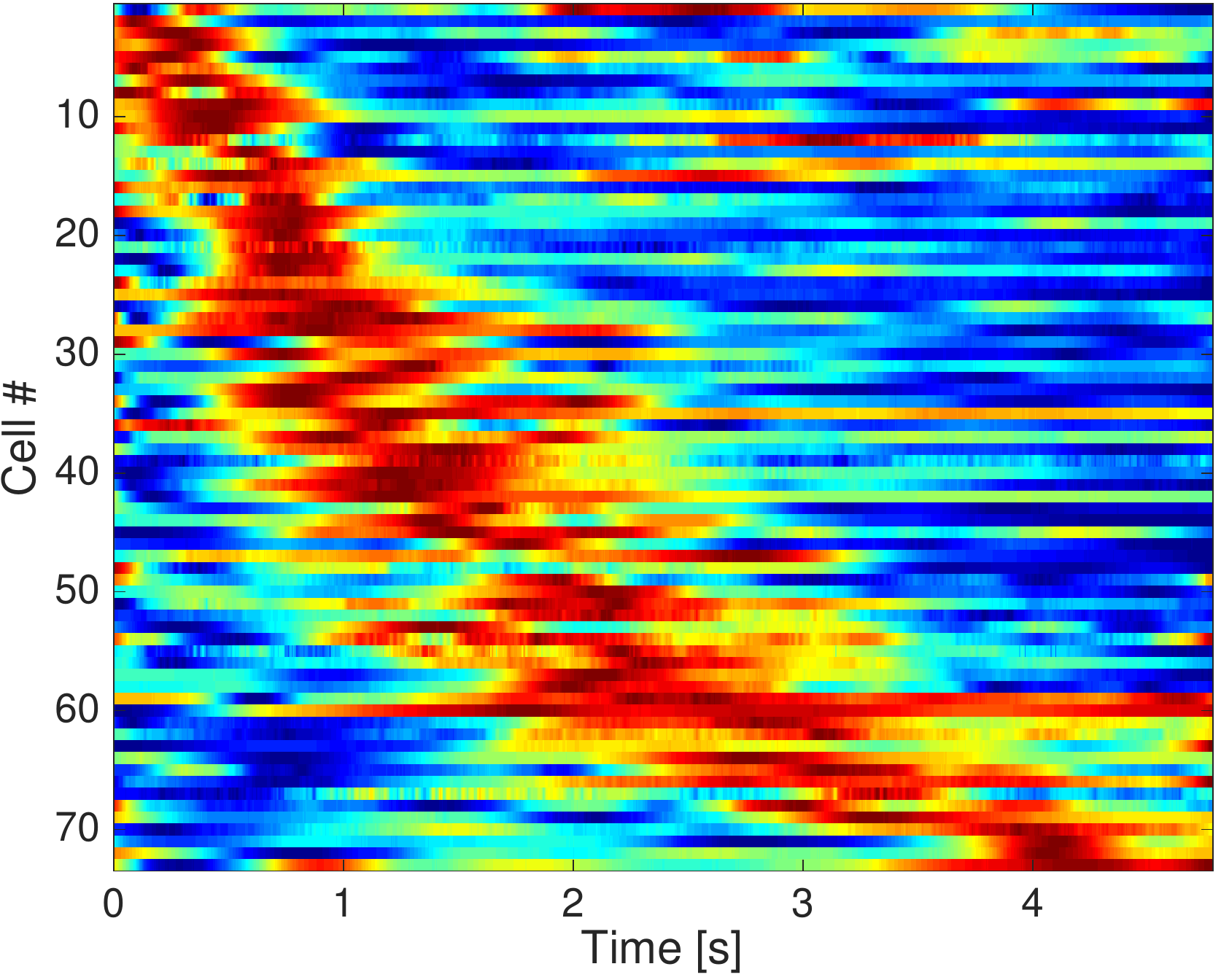}\\
		\textbf{b} &\\
	& 		\includegraphics[height=0.21\textheight]{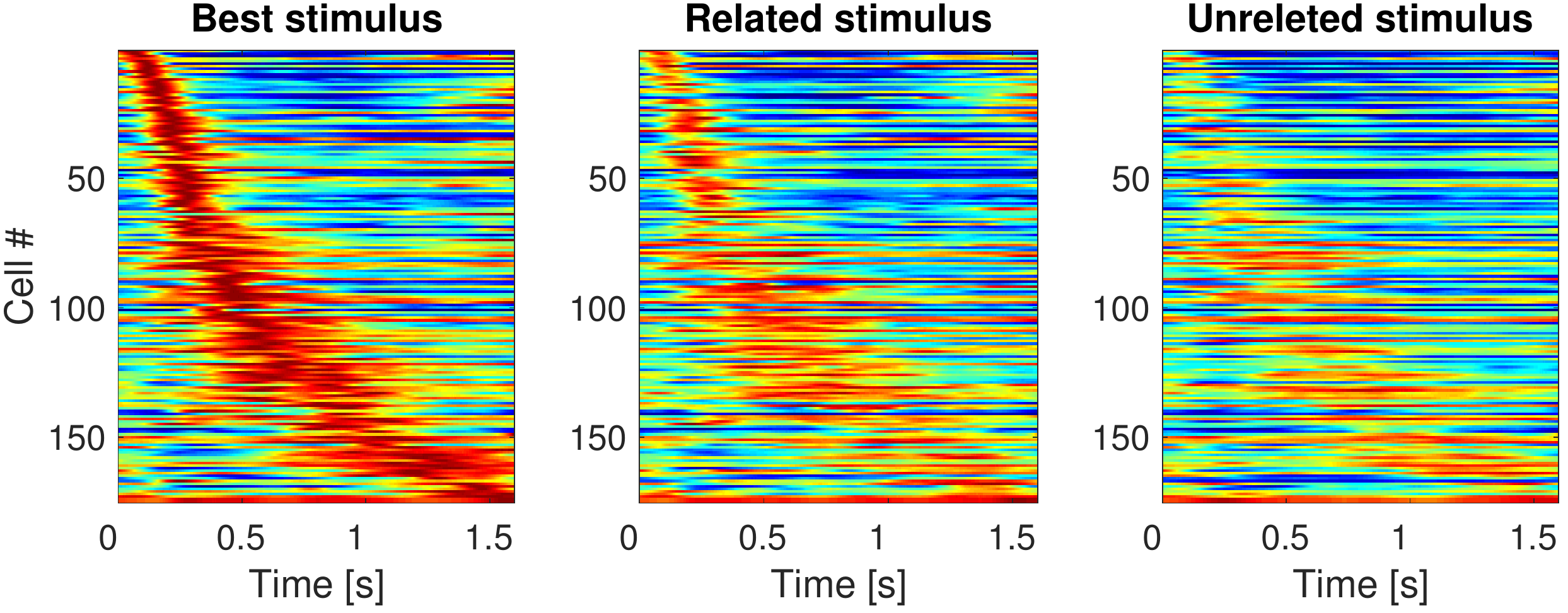}
\end{tabular}
\caption{Recordings of sequentially-activated time cells in different
		behavioral tasks, different mammalian species and different
		brain regions.  \textbf{a.}
		Sequences of time cells in the brain contain information about the
		time of past events. Each of the four plots shows the activity of many
		neurons during the delay period of a behavioral task.  In each plot
		each row gives the average firing rate as a function of time for one
		neuron.  Red colors indicate a high firing rate. Because neurons fire
		for a circumscribed period of time during the delay, these neurons
		could be used to decode the time at which the delay started.  Put
		another way, each neuron can be understood as coding for the presence
		of the start of the delay at a lag $\taustar$ in the past. The
		number of cells active at any one time decreases as the delay
		unfolds (note the curvature) and the firing fields spread (note the
		increasing width of the central ridge).  This reflects a decrease in
		accuracy for time as the start of the delay recedes further into the
		past.  From left to right: mouse mPFC during a spatial working memory
		task, after Bolkan et al., 2017; rat hippocampus, during the delay period
		of a working memory task, after MacDonald et al., 2011; rat mPFC
		during a delay period of temporal discrimination task, after Tiganj et
		al., 2016.
		\textbf{b.}
		Sequentially activated cells in monkey lPFC encode time conjunctively
		with stimulus identity during delayed-match-to-category task. Animals
		were presented with stimuli chosen from four categories
		(\textsc{dogs}, \textsc{cats}, \textsc{sports cars} and \textsc{sedan
		cars}). Based on visual similarity the stimuli belonged to two
		category sets (\textsc{animals} and \textsc{cars}). The time interval
		shown on the plots includes 0.6~s sample stimulus presentation and 1~s
		delay interval that followed the sample stimulus. Each of the three
		heatmaps shows the response of every unit classified as a time cell.
		The units show distinct firing rate for different stimuli that started
		the delay interval, reflecting the visual similarity (magnitude of the
		response for \textsc{Related stimulus} was larger than for
		\textsc{Unrelated stimulus}) and indicating stimulus selectivity
		of time cells. After Tiganj et al., 2018. 
\label{fig:heatmaps}
}
\end{figure*}

Figure~\ref{fig:f_and_M}a shows the sequential, spreading activation with
logarithmically spaced $\taustar$ for three different transient stimuli.  This
mathematical model for estimating the past has properties that resemble
sequentially activated time cells \cite<compare to Fig.~\ref{fig:heatmaps};
see also>{HowaEtal14,TigaEtal18a}.  Previous biophysical modeling has
developed  a neurally plausible mechanism for implementing leaky integrators
with a spectrum of time constants \cite{TigaEtal15} and for constructing  a
circuit implementing the inverse transform \cite{LiuEtal18}.  
 

\nocite{BolkEtal17}

$\ftildebold$  approximates $\mathbf{f}$ leading up to the present.  However,
the precision of the approximation decreases for events further in
the past.  One way to see this is that the duration over which 
$\ftilde_{\taustar,\alpha}$ is activated by a delta function input increases
as one chooses larger values of $|\taustar|$.   However, this inaccuracy is scale
invariant; the spread in time for a neuron with a particular $\taustar$ 
is a rescaled version of the firing of another neuron that received the same
input but has a different value of $\taustar$.  Put another way, the activity
of every neuron receiving a delta function input obeys the same time
dependence in units of $t/|\taustar|$.
This rescaling of the activity of neural response in time also has a
correspondence in the pattern of activity across neurons with different values
of $\taustar$ as the stimulus recedes into the past.  At any moment, when the
stimulus is $t_o$ time in the past, there is a bump of activity centered around the
neurons with $\taustar \simeq t_o$.   However, the difference in the value of
$\taustar$ between adjacent neurons is not constant (for instance note the
increasingly spread points in Figure~\ref{fig:f_and_M}b).
With logarithmic spacing of $\taustar$ values, the shape of the bump of
activity \emph{across
neuron number} remains of constant width as the stimulus recedes into the past
\cite{HowaEtal15}.


\subsection{Constructing an associative memory}
\label{sec:associative}
At each time $t$, an associative memory tensor $\mathbf{M}$ is updated
with the outer product of the current input state $\mathbf{f}$ and 
$\ftildebold$ (Figure~\ref{fig:f_and_M}b).
Hence $\mathbf{M}$ is a three-tensor. At each moment, $\mathbf{M}$ is updated
with the simple Hebbian learning rule:
\begin{equation}
		\frac{d{M}_{\taustar,\beta,\alpha}}{dt} = \lambda \
		{f}_\beta(t) \ \ftilde_{\taustar,\alpha}(t).
		\label{eq:dM}
\end{equation}
Here $\lambda$ is a
learning rate that we choose to be 1. $\mathbf{M}$ can be implemented 
as set of synaptic weights learned through Hebbian plasticity.  Because $\tilde{\mathbf{f}}_{\taustar,\alpha}$ stores a coarse-grained estimate of the past, its average over many experiences, $\mathbf{M}$, is a coarse-grained estimate of the lagged cooccurrence of each pair of states:
\begin{equation}
		M_{\taustar,\beta,\alpha} \propto
		P\left[
		\textbf{f}(t+|\taustar|)= \beta, \ \ \textbf{f}(t)=\alpha\right],
	\label{eq:prob}
\end{equation}
where $P$ denotes probability. 

We can also construct an estimate of the conditional probability by
normalizing $\mathbf{M}$ as follows:
\begin{equation}
		\bar{M}_{\taustar,\beta,\alpha} \equiv 
		\frac{ {M}_{\taustar,\beta,\alpha} }{
			\sum_{i \in I}  {M}_{\taustar,\beta,i}
		}.
		\label{eq:Mhat}
\end{equation}
One could imagine that this 
normalization is implemented on-line by a divisive presynaptic normalization mechanism
\cite{BeckEtal11}.  
Now $\bar{\mathbf{M}}$ is an associative memory that provides a coarse-grained
estimate of the conditional probability of state $\beta$ following state
$\alpha$ at a lag of $|\taustar|$:
\begin{equation}
		\bar{M}_{\taustar,\beta,\alpha}  \propto 
		P\left[\textbf{f}(t + |\taustar|)=\beta\ | \  \textbf{f}(t)= \alpha\right].
\label{eq:condprob}
\end{equation}
As we will see in the next subsection, by multiplying $\bar{\mathbf{M}}$  from the right with a
current state, we can generate the probability of all other states following
at each possible lag.

\subsection{Estimating a future timeline}
\label{sec:predict}
 $\mathbf{M}$ stores the pairwise temporal relationships between all
 stimuli subject to logarithmic compression.  
 At the moment a state is experienced, the  history leading up to that state is
  stored in $\mathbf{M}$ (eq.~\ref{eq:dM}).  After many presentations,
  $\mathbf{M}$ records the probability that each state is preceded by every
  other state at each possible lag.  This record of the past can also be used
  to predict the future.
  By multiplying $\bar{\mathbf{M}}$ with the current state from the right we can 
 generate an estimate of the future. 
In a general case, let us consider $\mathbf{f}(t)$ that can have multiple
stimuli presented at the same time. Stimuli that will follow the present input $\mathbf{f}(t)$ at a time lag $|\taustar|$ can be estimated from the information recorded in $\bar{\mathbf{M}}$:
 \begin{eqnarray}
		\mathbf{p}_{-\taustar} &\equiv& \bar{\mathbf{M}}_{\taustar} \mathbf{f}\\
		{p}_{-\taustar,\beta} &=& \sum_{i \in I}  \bar{M}_{\taustar,\beta,i}
				f_i.
		\label{eq:method1general}
\end{eqnarray}
 Like $\mathbf{F}$ and $\ftildebold$,
 $\mathbf{p}$ can be understood as a 2-D array indexed by
 stimulus identity and 
 $\taustar$. However, whereas for $\ftildebold$, $\taustar$ is negative
 corresponding to estimates of the past, for $\mathbf{p}$ the values of
 $\taustar$ are positive, corresponding to estimates of the future.   The value
 of $\taustar$ for the $i$th row of $\ftildebold$ and the value of $\taustar$
 for the $i$th row of $\mathbf{p}$ have the same magnitude but are opposite in sign. ${p}_{-\taustar,\beta}$ is a magnitude of the prediction that state $\beta$ will follow present input $\mathbf{f}$ at a time lag $|\taustar|$. When the input is interpretable as a probability density function (when $\sum|\mathbf{f}|=1$), then $\mathbf{p}_{-\tstr}$ is also a probability density function. When $\mathbf{f}$ is not a probability density function, $\mathbf{p}_{-\tstr}$ is not either. 

In a more specific case, when $\mathbf{f}(t)$ can have only one stimulus presented at the same time,  magnitude of the prediction that state $\beta$ will follow the present input, say state $\alpha$, at a time lag $|\taustar|$ is a scalar stored in $\bar{M}_{\taustar,\beta,\alpha}$:
\begin{eqnarray}
		{p}^\alpha_{-\taustar,\beta} &=& \bar{M}_{\taustar,\beta,\alpha}.
			\label{eq:method1}
\end{eqnarray}


Note that $\mathbf{p}^\alpha$ inherits the same compression
present in $\ftildebold$.  The ``blur'' in the estimate
of the time of presentation of a past stimulus in $\ftildebold$ with $\taustar
< 0$ naturally leads to an analogous blur in $\mathbf{p}^\alpha$ as a function of
future time $\taustar > 0$.  
Expected future outcome at a lag $\taustar$ can be estimated by examining the
states predicted at that lag and estimating the reward status of each.
Properties of this representation of future
time are illustrated in more detail in Section~\ref{sec:examples}.


\section{Illustrating the properties of the representation of future time}

\label{sec:examples}


In this section we illustrate properties of the representation of future time
constructed by multiplying $\bar{\mathbf{M}}$ with a particular state vector (Eq.~\ref{eq:method1}).
In subsection~\ref{sec:scale-inv} we demonstrate that the representation that
results is scale-invariant.
In subsection~\ref{sec:value} we show that a cached value for each state can
be computed, resulting in a scale-invariant  value that is discounted
according to a power law. 
In subsection~\ref{sec:tempfunc} we illustrate the flexibility of this method
in generating non-monotonic functions enabling the user to solve problems such
as the ``hot coffee'' problem described in the introduction.
In subsection~\ref{sec:trajectory}
we demonstrate that future time gives an estimate summed over all possible paths. Finally, in subsection~\ref{sec:DM} we demonstrate application of this approach in decision making. 



\subsection{Scale-invariance of future time}
\label{sec:scale-inv}
\begin{figure*}
\centering
\begin{tabular}{lc}
\textbf{a} & \\
&		\includegraphics[width=0.8\textwidth]{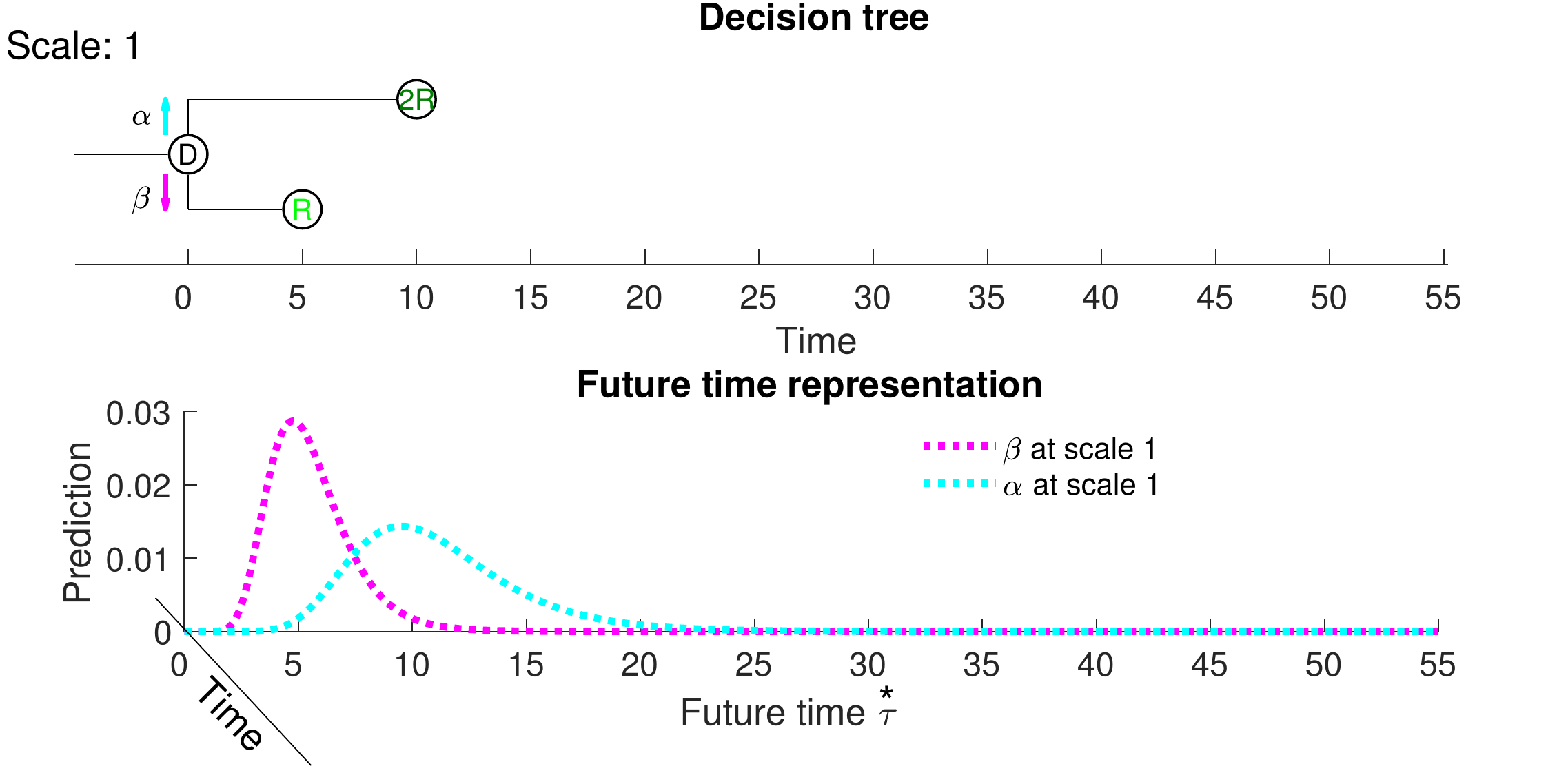}\\
\textbf{b} &\\
&		\includegraphics[width=0.8\textwidth]{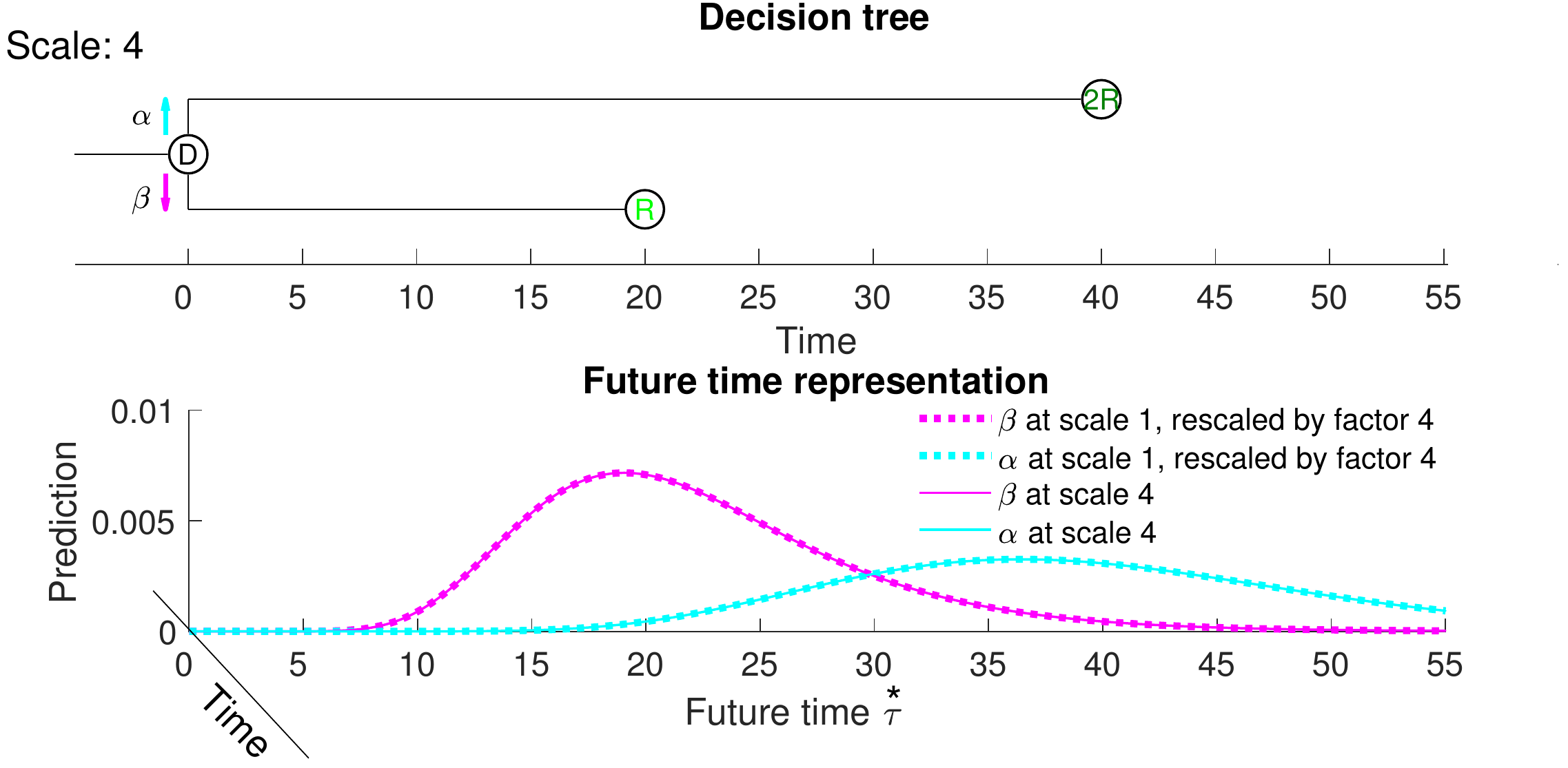}	
 \end{tabular}\\
 \caption{The representation of future time is scale-invariant. In this and
		 subsequent figures, the agent evaluates the degree to which each
		 of two states, $\alpha$ and $\beta$, predict a desired outcome
		 \textsc{r}.  The decision tree in the environment is shown in  the
		 top panel.  The estimate of the future cued by each of the two states
		 $\alpha$ and $\beta$ as a function of future time $\taustar$ is shown by
		 the lines at the bottom. 
		 \textbf{a.}  A simple decision tree in which $\alpha$ predicts reward
		 after 5 units of time and $\beta$ predicts reward after 10 units of
		 time.  The dashed lines show that the cues predict reward at
		 different times. Note that the prediction of events further in the
		 future is made with less precision.  (The reward predicted by
		 $\beta$ is twice the size of the reward
		 predicted by $\alpha$ to make the figure easier to read.)
		 \textbf{b.}  The same decision tree, but with the temporal intervals
		 rescaled by a factor of 4.  The solid lines show the predictions from
		 the environment in \textbf{a} rescaled by a factor of 4 (i.e.,
		 stretched by a factor of 4 and multiplied by 4).   Note that the
		 functions in the two environments are precisely rescaled versions of
		 one another.  
		 \label{fig:rescaling_time}
}  
\end{figure*}

If two environments differ only in their temporal scale, an artificial agent
based on a scale-invariant algorithm will take the same actions in both
environments. This property is illustrated for this method through a simple
toy example in Figure~\ref{fig:rescaling_time}. 
In this example, there are two states to choose from, $\alpha$ and $\beta$,
and a third rewarding state \textsc{r} that the agent is interested in
predicting.
The two environments shown in
Figure~\ref{fig:rescaling_time} differ only in the temporal spacing between
different stimuli. The bottom environment (marked as \textit{Scale 4}, Figure~\ref{fig:rescaling_time}b) is a
temporally stretched version of the top environment (marked as \textit{Scale
1}, Figure~\ref{fig:rescaling_time}a).  Stretching the time axis of the top environment by 4 times would give
exactly the bottom environment. At the decision point $D$ at time $t=0$ the
agent needs to choose either state $\alpha$ or state $\beta$ (the example is designed as a deterministic Markov decision process so taking an action can be understood as directly selecting a state).


Under the assumption that the agent has explored the environment by choosing
each direction at least once, all needed temporal associations are stored in
$\bar{\mathbf{M}}$. Next time when the agent faces the decision point at time
$t=0$ it can construct the future time as in Eq.~\ref{eq:method1}. The
predictions  $\mathbf{p}^\alpha$ and $\mathbf{p}^\beta$ constructed separately
for $\alpha$ and $\beta$ both give  power-law discounted estimates of the
expected future outcome that rescale with rescaling of the environment. 



\subsection{Computing cached power-law discounted stimulus value by
integrating over the timeline}
\label{sec:value}
There are circumstances where a decision-maker does not have time to evaluate
a compressed function over future time and a cached value of each state
would be sufficient.  A cached value can be computed by maintaining, for each
state, an average value over future time updated by taking an integral over
the future:
\begin{equation}
	V^\alpha(t) = \sum_{i \in I} r_{i}  \int_0^\infty  p^{\alpha}_{\taustar,i}\
	g_{\taustar}\ d\taustar,
	\label{eq:V}
\end{equation}
where $\mathbf{r}$ is a column vector describing the value of each state and
$g_{\taustar}$ is the number density of $\taustar$ values
$\frac{dN}{d\taustar}$. The number density $g_{\taustar} = dN/d\taustar$ specifies how many units are used to represent a particular spacing of $\taustar$. For instance, if spacing between $\taustar$ nodes would be linear the number density $g_{\taustar}$ would be 1. With logarithmic spacing of  $\taustar$ the number density goes down as $1/\taustar$.

\begin{figure*}
\centering
\begin{tabular}{lclclc}
\textbf{a} 
&&\textbf{b} 
&&\textbf{c} \\

&		\includegraphics[width=0.30\textwidth]{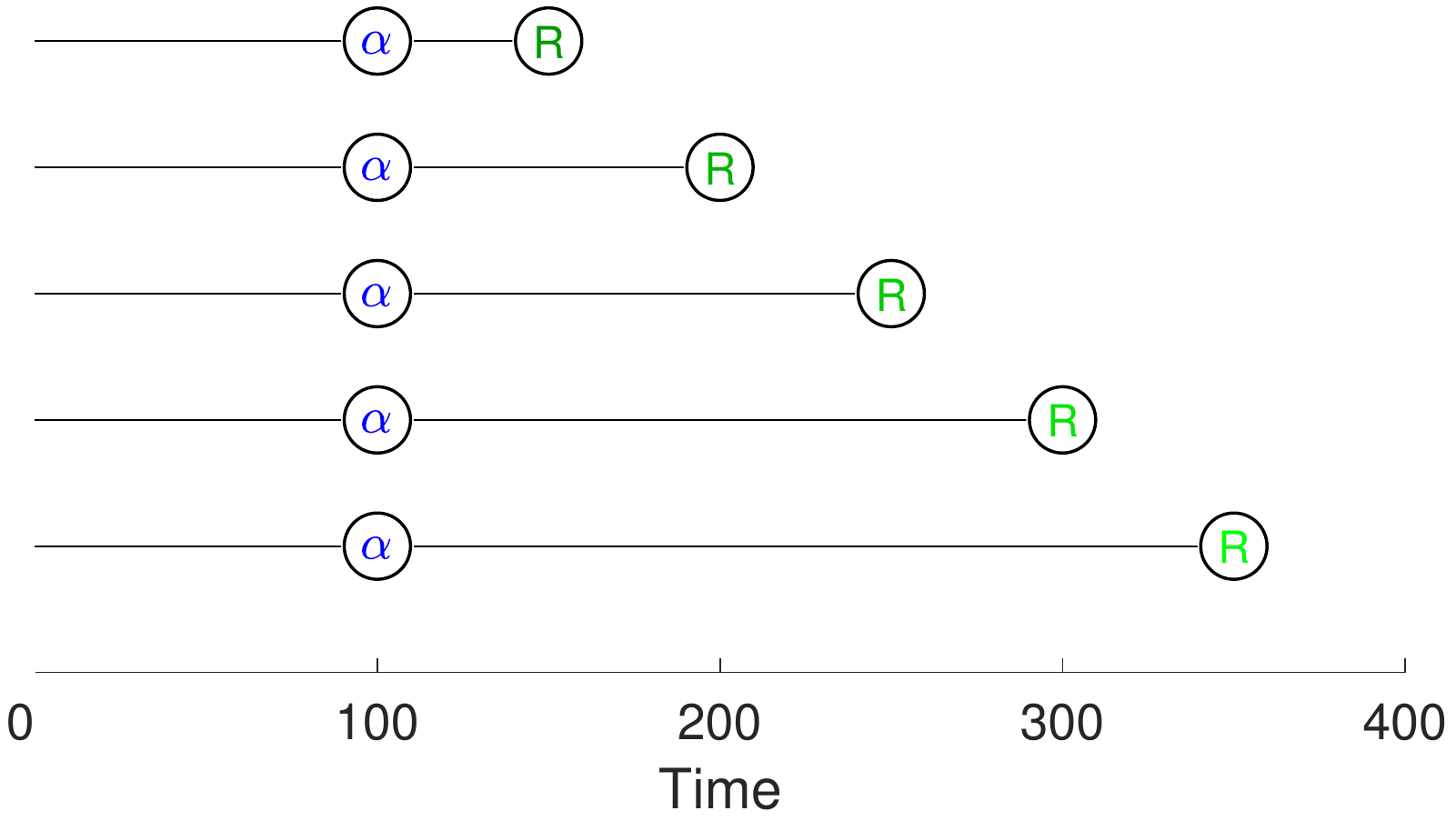}	
&&		\includegraphics[width=0.25\textwidth]{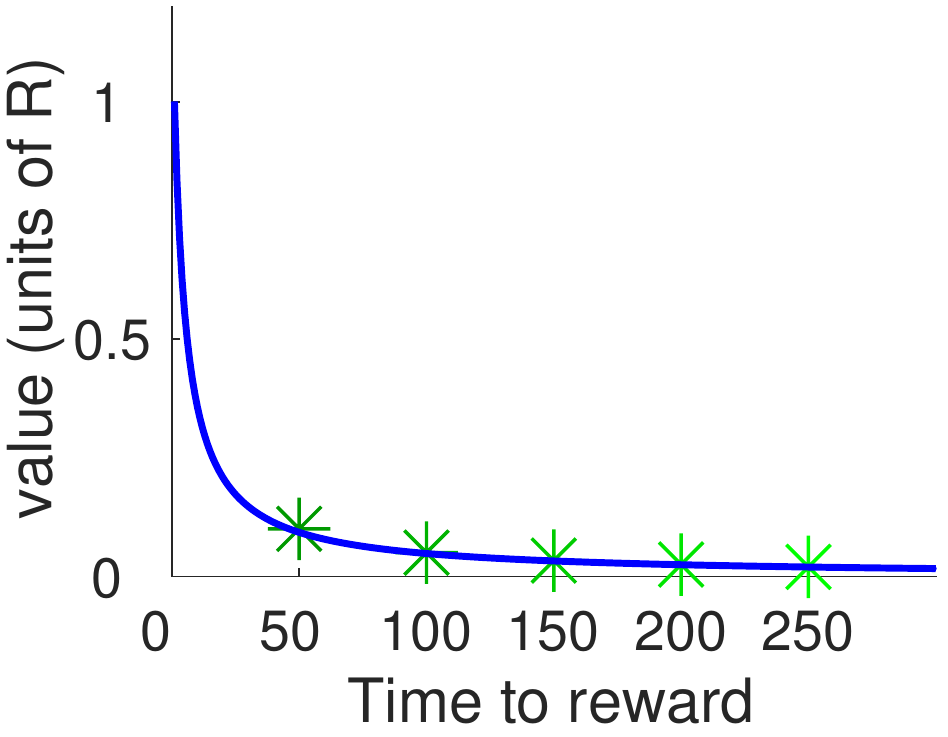}	
&&		\includegraphics[width=0.25\textwidth]{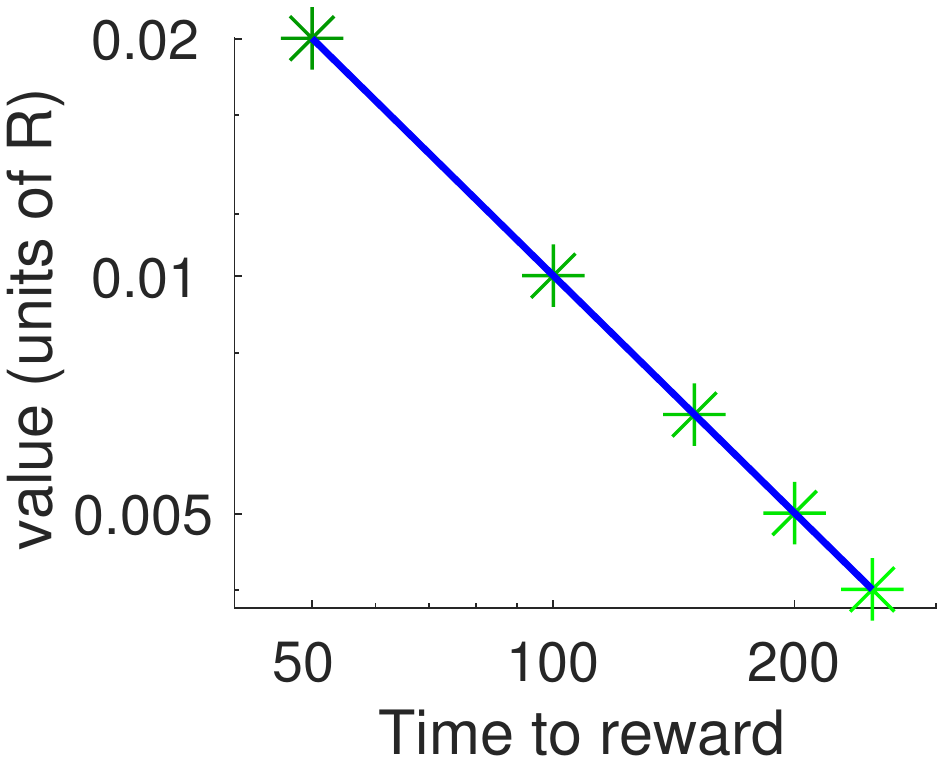}
 \end{tabular}
 \caption{ 
The value aggregated by integrating over future time obeys power-law
discounting.  {\textbf{a.} Constructing prediction
of the future reward. An agent observed a temporal sequence consisting of a
state $\alpha$, followed by a rewarding state \textsc{R} at some
delay. 
\textbf{b.}  The value of $\alpha$ computed according to Eq.~\ref{eq:V} as a
function of the delay between $\alpha$ and reward (expressed in the units of the reward $R$).  The value associated with
the five evenly spaced delays in \textbf{a} are shown as star symbols.  The
blue line is a power law with exponent $-1$.
\textbf{c.} Same as \textbf{b} but on  log-log axes. 
}
\label{fig:prediction_value}}
\end{figure*}

In order to ensure Weber-Fechner spacing, we here set
$g_{\taustar}={\taustar}^{-1}$, but one could in general augment this by
including a function to differentially weight the contribution of different
values of $\taustar$.  As long as that function does not introduce a scale,
the cached value computed in this way will remain scale invariant (power-law).
Figure~\ref{fig:prediction_value} illustrates properties of value computed
from Eq.~\ref{eq:V}. 

Applying Eq.~\ref{eq:V} to the example shown in Figure~\ref{fig:rescaling_time} reveals that the ratio of the values  for states $\alpha$ and $\beta$
is constant when time is rescaled.  This means that the relative values
assigned to various choices do not depend on the time-scale of the
environment, but only on their relative magnitude and timing.




\subsection{Non-monotonic functions over future time}
\label{sec:tempfunc}
\begin{figure*}
\centering
\includegraphics[width=0.7\textwidth]{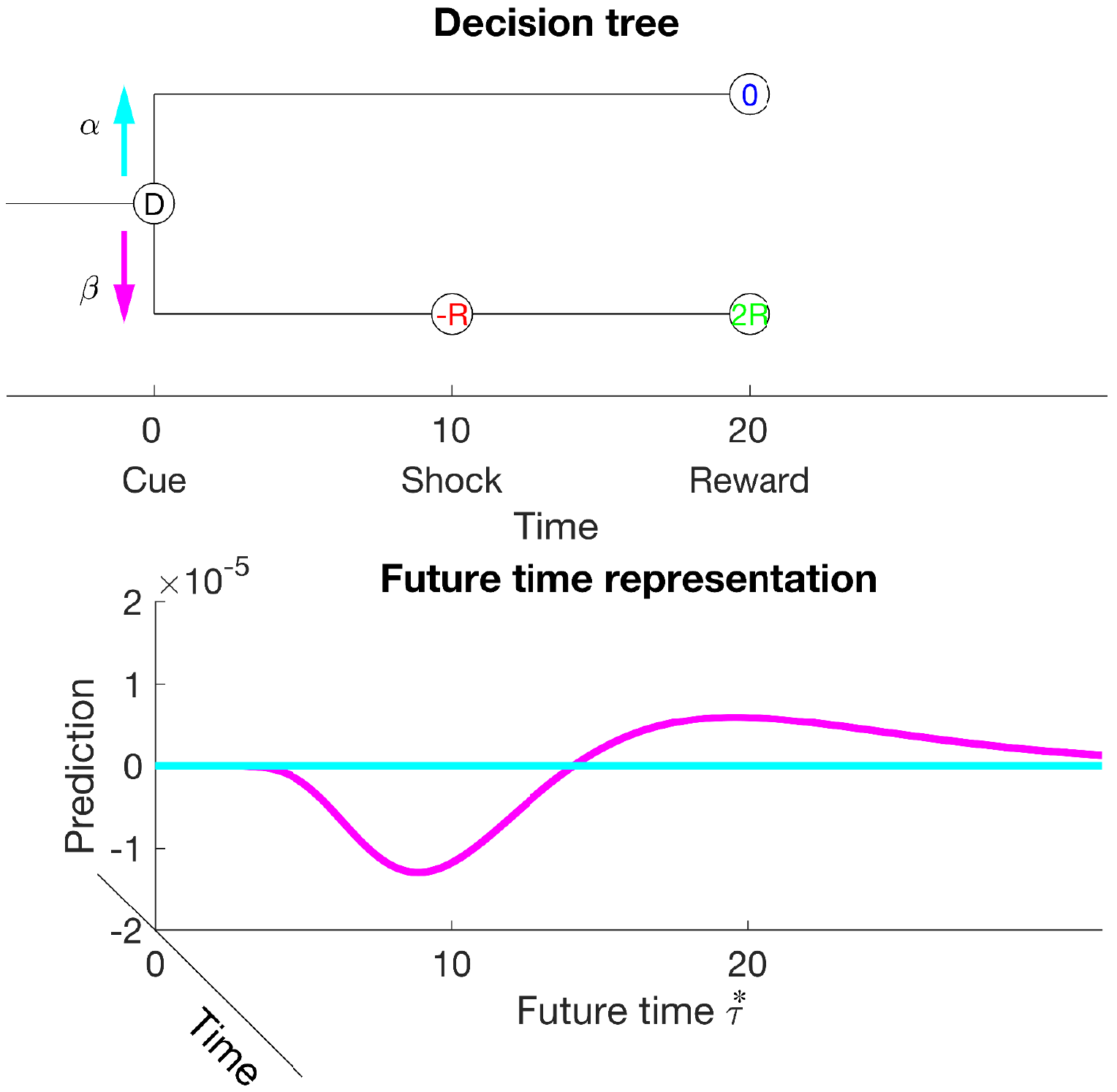}	
 \caption{
		 The estimate of a future timeline enables the decision-maker to
		 anticipate different outcomes at different points in time.
		 Top: Choice $\alpha$ is neutral, predicting neither reward nor
		 punishment. Choice $\beta$ results in a 
 negative outcome (e.g., a shock) after 10 units of time and then a large positive
 outcome after 20 units of time. 
 Bottom:  The representation of future time induced by each choice varies as a
 function of the temporal horizon.  $\alpha$ is preferable to $\beta$ at short
 delays but $\beta$ is preferable to $\alpha$ at longer delays.  A
 decision-maker could incorporate this information about the future time
 course when the choices are presented.
 \label{fig:future_time}}
\end{figure*}

In traditional RL, the value of each state is a scalar.
The approach introduced here provides a recipe for simulating a function of a
logarithmically compressed future.  The example in
Figure~\ref{fig:future_time} illustrates one case in which this type of
representation has an advantage over the scalar representation.  
In this example state $\alpha$ is neutral; no meaningful outcome follows it.
However, state $\beta$ is followed sequentially by  a negative 
outcome (e.g., a burned mouth)  and then later by a positive outcome 
(e.g., delicious coffee). 
The ability to simulate outcomes as  a function of future time can enable the
agent to make decisions in a more flexible way (by dynamically altering the time horizon of planning) than would be possible if all
the available information about the future was expressed as a scalar. 

Notice that the same amount of information is conveyed even when having only the set of exponentially decaying neurons ($\mathbf{F}$ neurons). However, applying the inverse Laplace transform and estimating the future  as proposed here allows the agent to examine the future directly in the units of time, without need for an additional decoder. This type of representations provides direct access to temporal order and distance. In general, coding  with bell-shaped turning curves appears to be widely used in the brain with place cells in the hippocampus, angle selective cells in visual and motor areas being some of well known examples.  

\subsection{Future time sums over trajectories}
\label{sec:trajectory}

\begin{figure*}
\centering
\includegraphics[width=0.7\textwidth]{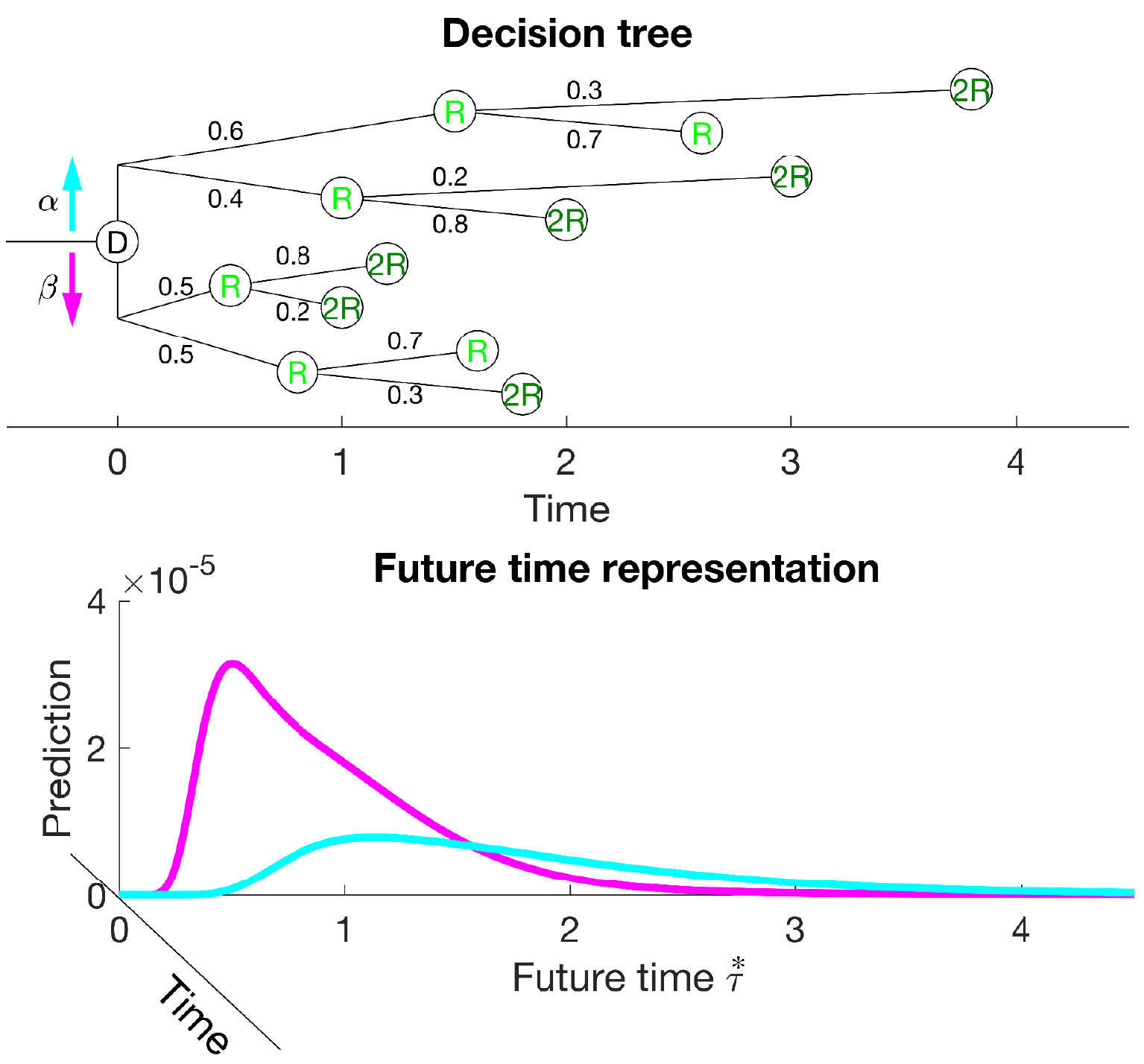}	
 \caption{The estimate of a future timeline sums over all possible future trajectories.
		Top: The decision tree for a complex choice.  
		 Both $\alpha$ and $\beta$ lead to  a complex set of possible outcomes that
		 occur at specific times with given probabilities (shown as numbers
		 near each branch of the tree). 
		 Bottom: The estimate of a future timeline averages over different
		 possible paths with each outcome weighted by its probability of
		 occurrence given the choice stimulus ($\alpha$ or $\beta$). 
\label{fig:sum_trajectories}}
\end{figure*}

Figure~\ref{fig:sum_trajectories} illustrates an important property of the
proposed approach: simulated future time  provides a probability of each
stimulus $\taustar$ in the future summed across all possible future
trajectories. 
Let us assume that the agent has sampled
the environment sufficiently many times to learn the transition probabilities
and the temporal dynamics of the environment, which are now stored in
$\bar{\mathbf{M}}$. Now computing the prediction
$\mathbf{p}^{\alpha}_{\taustar}$ as in Eq.~\ref{eq:method1}  provides an overall
estimate of the reward averaged across all the future trajectories. However, it
retains information about how far in the future those outcomes will be
obtained.  
This property allows a rapid evaluation of different decision trees.
Evaluating a particular sequence of outcomes that depend on sequential actions
would still require supplementing this representation with a more traditional
model-based approach.  Moreover, correctly learning the outcomes requires
sampling the entire tree, which may be much slower than TD-based learning in
an environment with Markov statistics. 

Notice that in many problems in  RL states that follow the present state often change in response to the action taken by the agent. For simplicity we are studying the Pavlovian case (similar to previous authors like \citeNP{SchuEtal97}). In the control setting, we would need to simultaneously estimate M and a policy, since these are coupled. We think the interplay between prediction and control are very important and we leave that to future work.

\subsection{Temporally flexible decision making}
\label{sec:DM}
\begin{figure*}
\centering
\includegraphics[width=0.7\textwidth]{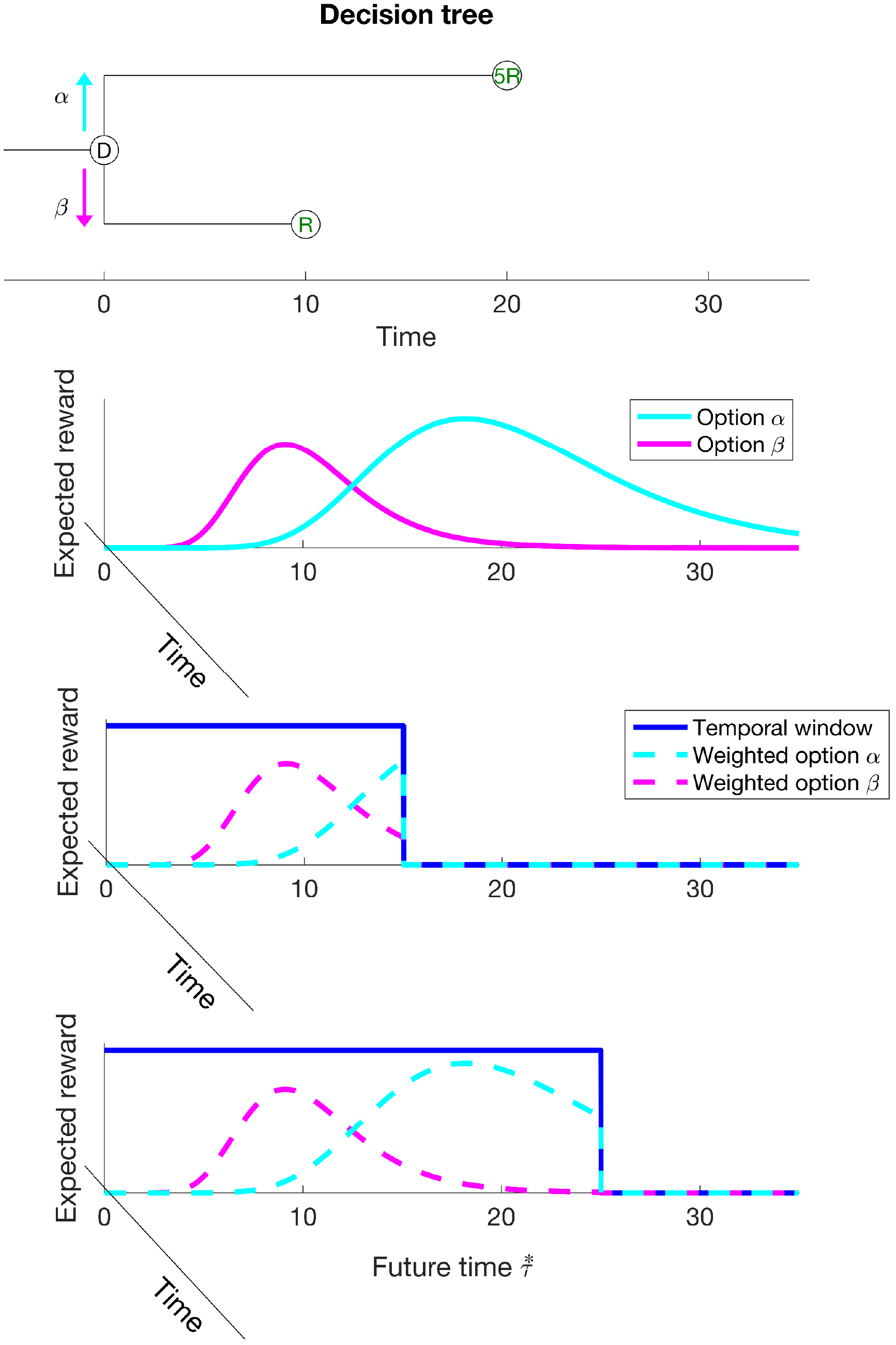}	
 \caption{Temporally flexible decision making. Consider an agent faced
		 with two options $\alpha$ and $\beta$ that differ in the time
		 course over which they predict reward (top).   
			Note that $\beta$ (magenta) predicts a larger reward, but
			further in the future, than does $\alpha$ (cyan). 
		 Representing a
		 function  over future time enables the agent to make decisions that
		 incorporate information about the value of the outcome to the
		 agent as a function of time.  
		For instance, under some circumstances, an agent might have more or
		less time to exploit a food source before some other pressing
		engagement. 	 The bottom two panels illustrate the value
		computation with each of two temporal windows.
		 In the middle panel, the temporal
		 window over which the agent can exploit the reward is narrow and the
		 agent chooses $\alpha$.  In the bottom panel, the temporal window
		 extends further into the future and the agent chooses $\beta$.
 \label{fig:DM}}
\end{figure*}

The ability to construct a timeline of the future events enables flexible decision
making that incorporates the decision-maker's temporal constraints.
For instance, consider making a decision about what to get for lunch while
waiting for a train.  The food option one pursues may be very different if one
has 15~minutes before the train arrives than if one has an hour before the
train arrives.  Because the model carries separate information about
\emph{when} outcomes will be available as well as their identity it is
possible to make decisions that differentially weight outcomes at different
points in the future.  If the decision-maker has a temporal window over which
outcomes are valuable, $w_{\taustar}$, then one can readily compute value using
a generalization of Eq.~\ref{eq:V}:
\begin{equation}
	V^\alpha(t) = \sum_{i \in I} r_{i} \int_0^\infty p^{\alpha}_{\taustar,i}w_{\taustar}\
	g_{\taustar}\ d\taustar,
	\label{eq:window}
\end{equation}
Figure~\ref{fig:DM} illustrates this capability.  In this
example, the model is presented with two alternatives that predict a valuable
outcome but with different magnitude and different time course. 
When the decision-maker approaches the choice with a narrow temporal window,
as in the case where the train will arrive in 15~minutes, choice \textsc{a} 
is more valuable.  However, when choosing using a broader temporal window, as in
the case where the train will arrive in one hour, choice \textsc{b} is more
valuable. 

A temporal representation of the future enables not only decision-making with
different temporal horizons, but also decision-making based on relatively
complex temporal demands.  Consider the case where an outcome is not valuable
in the immediate future, but only becomes valuable after some time has
passed---for instance perhaps one is not hungry now but will be hungry in one
hour.  These capabilities are comparable to those offered by model-based RL.
However, as discussed above, the representation of the future is
scale-invariant and can be computed rapidly.

\section{Behavioral and neural predictions}
\label{sec:predictions}
Earlier sections presented a method for constructing a compressed
estimate of the future. Because this approach is novel, there is not yet
empirical data to definitively evaluate key predictions of this approach. In
this section we describe neural and behavioral predictions and describe how
those could be tested experimentally. We also point to recent empirical
results, both behavioral and neural, that support the proposed hypothesis,
albeit obliquely.

\subsection{Cognitive scanning of the future} 
This paper proposes a neural mechanism for constructing a compressed
representation of the future.
In the cognitive psychology of working memory, prior findings from the
short-term judgment of recency (JOR) task suggest that people can scan a
compressed representation of the past.  For instance, \citeA{Hack80} presented
participants a series of letters rapidly and asked them to evaluate which of
two probes was experienced more recently.  The critical finding was that the
time to choose a probe depended on how far in the past that probe was
presented and did not depend on the recency of the other probe. These
findings suggested that participants sequentially examine a
temporally-organized representation of the past and terminate the search when
they find a target \cite<see also>{Mute79,Hock84,McElDosh93}. Furthermore, the time to choose a probe grew sublinearly with how far in the past the probe item was, suggesting that the temporally-organized memory representation is compressed (the results were consistent with the hypothesis discussed here that the compression is logarithmic). 
These findings from the memory literature suggest that an analogous procedure
could be used to query participants' expectations about the future.  
By setting the temporal windowing functions (eq.~\ref{eq:window}) to direct
attention to sequentially more distant points in the future, one could
sequentially examine an ordered representation of the past.

In order to evaluate whether human participants can scan across a
compressed temporally-ordered representation of the future,
\citeA{SingHowa17a} trained participants on a probabilistic sequence of
letters. After training, the sequence was occasionally interrupted with 
probes consisting of two letters. Participants were instructed to select the 
probe that is more likely to appear sooner. If the participants sequentially
scan a log-compressed timeline of future events then this predicts a 
pattern of results analogous to the findings from the JOR task. Specifically, evidence for sequential scanning would be that  
the response time in correct trials  depends only on
the lag of the more imminent probe.
Furthermore, in trials in which participants make an error,
the response time should depend on the lag of the less  imminent probe (this
is because if participants have missed the more imminent probe during the scanning
process, they will continue scanning until they reach the less imminent
probe). Evidence for compression of the temporally-ordered representation of the future would be a sublinear growth of response time with the lag to the probe that is selected. These predictions were confirmed \cite<see Figure 2b,>{SingHowa17a}.

\subsection{Neural signature of the compressed timeline of future events }
As discussed in the introduction, there is  ample evidence that neurons in the
mammalian brain can be used to decode what happened when in the past
\cite<e.g.,
Figure~\ref{fig:heatmaps}a,>{BolkEtal17,MacDEtal11,TigaEtal18a,SalzEtal16}.
By analogy, the present model  predicts that it should be possible to measure
neurons that predict what \emph{will} happen when in the future.  Because
predictions of the future cannot be dissociated from the past, it is possible
to have the same future predicted by distinct past events.
Consider a situation in which participants are trained on two distinct
sequences \textsc{a, b, c} and \textsc{x, y, c} and we record after training
from a region of the brain representing the future as described by
Eq.~\ref{eq:method1}. The model predicts that a common population of neurons
(coding for \textsc{c} two steps in the future)
should activate when either \textsc{a} or \textsc{x} are presented.
The response to the probe stimuli prior to learning of the sequences serves as
a control.  Similarly, a distinct population (coding for \textsc{c} one step
in the future) will be activated when either \textsc{b} or \textsc{y} are
presented.  In analogy to sequences of firing triggered by past
events \cite{TigaEtal18a}, this outcome would imply that similar sequences
of neural firing \emph{anticipate} similar outcomes (Figure~\ref{fig:heatmaps}b).


\section{Discussion}
In this paper we show that, given a compressed representation of the past, a
simple associative mechanism is sufficient to enable one to generate a
compressed representation of the future.  
A compressed representation of the past has been extensively observed in the
brain in many brain regions \cite{MacDEtal11,JinEtal09,TigaEtal18a}.
The associative mechanism we utilize can be understood as simple Hebbian
association.  The representation that is generated by this method has many
potentially desirable computational properties.

Because the representations of the past and the future are both
scale-invariant, it is not necessary to have a strong prior belief about the
relevant time scale of the problem one is trying to solve. A scale-invariant
learning agent ought to be able to solve problems in a wide range of learning
environments.  While it remains to be shown that the form of compression of
temporal sequences in the brain is \emph{quantitatively} scale-invariant
(rather than merely compressed), scale-invariance is a design goal that can be
implemented in artificial systems.

Because the method directly estimates a function over future states, rather
than an integral over future states, decision-makers can make adaptive
decisions that take into account the time of future outcomes.  The future
timeline constructed using this method differs from  traditional model-based
approaches.  After the association matrix $\mat{M}$ has been learned, the
computation of the future trajectory is computationally efficient and can be
accomplished in one parallel operation. $\mat{M}$ can be learned rapidly, allowing even one-shot learning, unlike, for instance, approaches based on backpropagation. In addition, the logarithmic form for
the future means that even if the decision-maker queries the representation
sequentially, the amount of time to access a future event goes up sublinearly. 
Recent behavioral evidence from human subjects shows just this result
\cite{SingHowa17a}.

Because the method treats time as a continuous variable, there is no need to
discretize time.  That is, the ``distance'' between two states need not be
filled with other states.  In TD learning, error propogates backward from one
state to a preceding state \emph{via} a gradient along intervening states.
Using the method in this paper, one can learn that \textsc{a} predicts
\textsc{b} separated by, say, 17.4~s without having to define a set of discrete
states that intervene.  The number of presentations necessary to establish a
relationship between two stimuli in \mat{M} depends on their number of pairings
rather than the lag that intervenes between them.


%
%
%

\subsection{Relationship to the successor representation}

The idea of efficiently computing compressed summaries of the future arises in
another approach to RL, based on the successor representation (SR;
\citeNP{Daya93}). Instead of estimating cached values (as in model-free
approaches) or transition functions (as in model-based approaches), the SR
estimates the discounted expected future occupancy of each state from every
other state. The SR can then be combined with an estimated reward function to
produce value estimates. Thus, this approach permits the computation of values
without expensive tree search or dynamic programming, but retains some of the
flexibility of model-based approaches by factoring the value function into
predictive and reward components. From a neurobiological and psychological
point of view, several lines of evidence have suggested that the brain might
use such a representation to solve RL problems \cite{MomeEtal17,StacEtal16}.

The SR has many interesting computational properties, but it still runs afoul
of the issues raised in this paper. In particular, the SR assumes exponential
discounting and consequently imposes a time scale. If the world obeys a Markov
process at the assumed time scale, then the SR will be able to efficiently
solve RL problems. However, as we pointed out, realistic environments consist
of problems occurring at many different scales.  Moreover, effective
decision-making requires explicit information about the time at which stimuli
are expected to occur.  Thus effective RL in the real world may require more
temporal flexibility than what the SR can provide.

\subsection{Relationship to models of episodic memory and planning}


RL models have long utilized a rich interplay between planning, action
selection, and  prediction of future outcomes  \cite<e.g.,>{SuttEtal98}.
\citeA{GersDaw17}, building on an earlier proposal by \citeA{LengDaya07},
proposed that retrieval of specific instances from memory could enhance
RL-based decision-making.   In psychology, the ability to consciously retrieve
specific instances from one's life is referred to as episodic memory
\cite{Tulv83}.  Episodic memory could enhance the capabilities of RL-based
models by enabling single-trial learning and bridging across multiple
experiences  with the same stimulus to discover relationships among
temporally-remote stimuli \cite{BunsEich96,ColeEtal95,WimmShoh12}.

Episodic memory has also been proposed to share a neural substrate with what
is referred to as ``episodic future thinking''  \cite{Tulv85a,SchaEtal07}.
Recovery of an episodic memory results in vivid recall of one's past self in a
particular spatiotemporal context different from one's present circumstances.
Episodic future thinking is defined as imagination of one's future self in a
circumstances different from the present.   Notably, behavioral and
neuroimaging work shows that amnesia patients who are impaired at episodic
memory also show deficits in episodic future thinking and that the brain
regions engaged by episodic memory performance overlap with the regions
engaged by episodic future thinking \cite{AddiEtal07,HassEtal07a,PaloEtal15}.

The present approach suggests the first steps towards  a computational bridge
between episodic memory for the past and planning based on future time.  In
this paper, we showed that a temporal history $\ftildebold$ can be used
to generate a prediction of the future \emph{via} an associative memory.
The sequentially-activated neurons predicted by $\ftildebold$ strongly
resemble sequentially-activated ``time cells'' measured in the hippocampus
\cite{MacDEtal11,PastEtal08}, a brain region implicated in episodic memory.
Moreover, the present approach is closely related to the temporal context
model, a computational approach that has been applied to behavioral results 
in a range of episodic memory paradigms
\cite<TCM,>{HowaKaha02a,SedeEtal08,PolyEtal09,GersEtal12}.  In TCM, items are
bound to the prevailing temporal context present when the item appeared
\emph{via} an associative context-to-item matrix.  The temporal history
$\ftildebold$ plays a role very similar to temporal context in TCM,
although in TCM, temporal context is an exponentially-weighted sum over
recent experience that introduces a scale rather than the scale-invariant
representation of the past $\ftildebold$.

The major departure of the present model from TCM is that we have not enabled
recovery of a previous history by an item and used to cue
future outcomes.  That is, one might imagine a model in which, rather than
cueing $\mathbf{M}$ with a particular state $\alpha$, one enables state
$\alpha$ to recover a previous state of $\ftildebold$ that preceded
$\alpha$ and then use that recovered temporal history to predict future
outcomes.  This kind of mechanism not only enables TCM to account for the
contiguity effect in episodic memory, but also  allows flexible learning
across similar events \cite{HowaEtal05}. Future work should explore to what
extent a similar contextual reinstatement process, instead in this case
reinstating the compressed scale-free representation of the past
\cite{HowaEtal15}, would help speed up learning or transfer of knowledge and
predictions as an agent explores a novel world in similar, but not identical,
trajectories \cite{Gers17}.

\section{Acknowledgments} We gratefully acknowledge  discussions with Karthik
Shankar and Ida Momennejad. This work was supported by NIBIB R01EB022864, NIMH
R01MH112169, NIH R01- 1207833, MURI N00014-16-1-2832 and NSF IIS 1631460.

\bibliography{bibdesk_zoran,bibdesk,bibdesk_marc} 

\begin{thebibliography}{}

\bibitem [\protect \citeauthoryear {%
Addis%
, Wong%
\BCBL {}\ \BBA {} Schacter%
}{%
Addis%
\ \protect \BOthers {.}}{%
{\protect \APACyear {2007}}%
}]{%
AddiEtal07}
\APACinsertmetastar {%
AddiEtal07}%
\begin{APACrefauthors}%
Addis, D\BPBI R.%
, Wong, A\BPBI T.%
\BCBL {}\ \BBA {} Schacter, D\BPBI L.%
\end{APACrefauthors}%
\unskip\
\newblock
\APACrefYearMonthDay{2007}{}{}.
\newblock
{\BBOQ}\APACrefatitle {Remembering the past and imagining the future: Common
  and distinct neural substrates during event construction and elaboration}
  {Remembering the past and imagining the future: Common and distinct neural
  substrates during event construction and elaboration}.{\BBCQ}
\newblock
\APACjournalVolNumPages{Neuropsychologia}{45}{7}{1363--1377}.
\PrintBackRefs{\CurrentBib}

\bibitem [\protect \citeauthoryear {%
Adler%
\ \protect \BOthers {.}}{%
Adler%
\ \protect \BOthers {.}}{%
{\protect \APACyear {2012}}%
}]{%
AdleEtal12}
\APACinsertmetastar {%
AdleEtal12}%
\begin{APACrefauthors}%
Adler, A.%
, Katabi, S.%
, Finkes, I.%
, Israel, Z.%
, Prut, Y.%
\BCBL {}\ \BBA {} Bergman, H.%
\end{APACrefauthors}%
\unskip\
\newblock
\APACrefYearMonthDay{2012}{}{}.
\newblock
{\BBOQ}\APACrefatitle {Temporal convergence of dynamic cell assemblies in the
  striato-pallidal network} {Temporal convergence of dynamic cell assemblies in
  the striato-pallidal network}.{\BBCQ}
\newblock
\APACjournalVolNumPages{Journal of Neuroscience}{32}{7}{2473-84}.
\newblock
\begin{APACrefDOI} \doi{10.1523/JNEUROSCI.4830-11.2012} \end{APACrefDOI}
\PrintBackRefs{\CurrentBib}

\bibitem [\protect \citeauthoryear {%
Akhlaghpour%
\ \protect \BOthers {.}}{%
Akhlaghpour%
\ \protect \BOthers {.}}{%
{\protect \APACyear {2016}}%
}]{%
AkhlEtal16}
\APACinsertmetastar {%
AkhlEtal16}%
\begin{APACrefauthors}%
Akhlaghpour, H.%
, Wiskerke, J.%
, Choi, J\BPBI Y.%
, Taliaferro, J\BPBI P.%
, Au, J.%
\BCBL {}\ \BBA {} Witten, I.%
\end{APACrefauthors}%
\unskip\
\newblock
\APACrefYearMonthDay{2016}{}{}.
\newblock
{\BBOQ}\APACrefatitle {Dissociated sequential activity and stimulus encoding in
  the dorsomedial striatum during spatial working memory} {Dissociated
  sequential activity and stimulus encoding in the dorsomedial striatum during
  spatial working memory}.{\BBCQ}
\newblock
\APACjournalVolNumPages{eLife}{5}{}{e19507}.
\PrintBackRefs{\CurrentBib}

\bibitem [\protect \citeauthoryear {%
Balsam%
\ \BBA {} Gallistel%
}{%
Balsam%
\ \BBA {} Gallistel%
}{%
{\protect \APACyear {2009}}%
}]{%
BalsGall09}
\APACinsertmetastar {%
BalsGall09}%
\begin{APACrefauthors}%
Balsam, P\BPBI D.%
\BCBT {}\ \BBA {} Gallistel, C\BPBI R.%
\end{APACrefauthors}%
\unskip\
\newblock
\APACrefYearMonthDay{2009}{}{}.
\newblock
{\BBOQ}\APACrefatitle {Temporal maps and informativeness in associative
  learning.} {Temporal maps and informativeness in associative
  learning.}{\BBCQ}
\newblock
\APACjournalVolNumPages{Trends in Neuroscience}{32}{2}{73--78}.
\PrintBackRefs{\CurrentBib}

\bibitem [\protect \citeauthoryear {%
Beck%
, Latham%
\BCBL {}\ \BBA {} Pouget%
}{%
Beck%
\ \protect \BOthers {.}}{%
{\protect \APACyear {2011}}%
}]{%
BeckEtal11}
\APACinsertmetastar {%
BeckEtal11}%
\begin{APACrefauthors}%
Beck, J\BPBI M.%
, Latham, P\BPBI E.%
\BCBL {}\ \BBA {} Pouget, A.%
\end{APACrefauthors}%
\unskip\
\newblock
\APACrefYearMonthDay{2011}{}{}.
\newblock
{\BBOQ}\APACrefatitle {Marginalization in neural circuits with divisive
  normalization} {Marginalization in neural circuits with divisive
  normalization}.{\BBCQ}
\newblock
\APACjournalVolNumPages{Journal of Neuroscience}{31}{43}{15310--15319}.
\PrintBackRefs{\CurrentBib}

\bibitem [\protect \citeauthoryear {%
Bliss%
\ \BBA {} Collingridge%
}{%
Bliss%
\ \BBA {} Collingridge%
}{%
{\protect \APACyear {1993}}%
}]{%
BlisColl93}
\APACinsertmetastar {%
BlisColl93}%
\begin{APACrefauthors}%
Bliss, T\BPBI V.%
\BCBT {}\ \BBA {} Collingridge, G\BPBI L.%
\end{APACrefauthors}%
\unskip\
\newblock
\APACrefYearMonthDay{1993}{}{}.
\newblock
{\BBOQ}\APACrefatitle {A synaptic model of memory: long-term potentiation in
  the hippocampus} {A synaptic model of memory: long-term potentiation in the
  hippocampus}.{\BBCQ}
\newblock
\APACjournalVolNumPages{Nature}{361}{6407}{31}.
\PrintBackRefs{\CurrentBib}

\bibitem [\protect \citeauthoryear {%
Bolkan%
\ \protect \BOthers {.}}{%
Bolkan%
\ \protect \BOthers {.}}{%
{\protect \APACyear {2017}}%
}]{%
BolkEtal17}
\APACinsertmetastar {%
BolkEtal17}%
\begin{APACrefauthors}%
Bolkan, S\BPBI S.%
, Stujenske, J\BPBI M.%
, Parnaudeau, S.%
, Spellman, T\BPBI J.%
, Rauffenbart, C.%
, Abbas, A\BPBI I.%
\BDBL {}Kellendonk, C.%
\end{APACrefauthors}%
\unskip\
\newblock
\APACrefYearMonthDay{2017}{}{}.
\newblock
{\BBOQ}\APACrefatitle {Thalamic projections sustain prefrontal activity during
  working memory maintenance} {Thalamic projections sustain prefrontal activity
  during working memory maintenance}.{\BBCQ}
\newblock
\APACjournalVolNumPages{Nature Neuroscience}{20}{7}{987--996}.
\PrintBackRefs{\CurrentBib}

\bibitem [\protect \citeauthoryear {%
Brown%
, Steyvers%
\BCBL {}\ \BBA {} Hemmer%
}{%
Brown%
\ \protect \BOthers {.}}{%
{\protect \APACyear {2007}}%
}]{%
BrowEtal07}
\APACinsertmetastar {%
BrowEtal07}%
\begin{APACrefauthors}%
Brown, S.%
, Steyvers, M.%
\BCBL {}\ \BBA {} Hemmer, P.%
\end{APACrefauthors}%
\unskip\
\newblock
\APACrefYearMonthDay{2007}{}{}.
\newblock
{\BBOQ}\APACrefatitle {Modeling experimentally induced strategy shifts.}
  {Modeling experimentally induced strategy shifts.}{\BBCQ}
\newblock
\APACjournalVolNumPages{Psychological Science}{18}{1}{40-5}.
\PrintBackRefs{\CurrentBib}

\bibitem [\protect \citeauthoryear {%
Bunsey%
\ \BBA {} Eichenbaum%
}{%
Bunsey%
\ \BBA {} Eichenbaum%
}{%
{\protect \APACyear {1996}}%
}]{%
BunsEich96}
\APACinsertmetastar {%
BunsEich96}%
\begin{APACrefauthors}%
Bunsey, M.%
\BCBT {}\ \BBA {} Eichenbaum, H\BPBI B.%
\end{APACrefauthors}%
\unskip\
\newblock
\APACrefYearMonthDay{1996}{}{}.
\newblock
{\BBOQ}\APACrefatitle {Conservation of hippocampal memory function in rats and
  humans} {Conservation of hippocampal memory function in rats and
  humans}.{\BBCQ}
\newblock
\APACjournalVolNumPages{Nature}{379}{6562}{255-257}.
\PrintBackRefs{\CurrentBib}

\bibitem [\protect \citeauthoryear {%
Cole%
, Barnet%
\BCBL {}\ \BBA {} Miller%
}{%
Cole%
\ \protect \BOthers {.}}{%
{\protect \APACyear {1995}}%
}]{%
ColeEtal95}
\APACinsertmetastar {%
ColeEtal95}%
\begin{APACrefauthors}%
Cole, R\BPBI P.%
, Barnet, R\BPBI C.%
\BCBL {}\ \BBA {} Miller, R\BPBI R.%
\end{APACrefauthors}%
\unskip\
\newblock
\APACrefYearMonthDay{1995}{}{}.
\newblock
{\BBOQ}\APACrefatitle {Temporal Encoding in Trace Conditioning} {Temporal
  encoding in trace conditioning}.{\BBCQ}
\newblock
\APACjournalVolNumPages{Animal Learning \& Behavior}{23}{2}{144-153}.
\PrintBackRefs{\CurrentBib}

\bibitem [\protect \citeauthoryear {%
Daw%
\ \BBA {} Dayan%
}{%
Daw%
\ \BBA {} Dayan%
}{%
{\protect \APACyear {2014}}%
}]{%
DawDaya14}
\APACinsertmetastar {%
DawDaya14}%
\begin{APACrefauthors}%
Daw, N\BPBI D.%
\BCBT {}\ \BBA {} Dayan, P.%
\end{APACrefauthors}%
\unskip\
\newblock
\APACrefYearMonthDay{2014}{}{}.
\newblock
{\BBOQ}\APACrefatitle {The algorithmic anatomy of model-based evaluation} {The
  algorithmic anatomy of model-based evaluation}.{\BBCQ}
\newblock
\APACjournalVolNumPages{Philosophical Transactions of the Royal Society B:
  Biological Sciences}{369}{1655}{}.
\newblock
\begin{APACrefDOI} \doi{10.1098/rstb.2013.0478} \end{APACrefDOI}
\PrintBackRefs{\CurrentBib}

\bibitem [\protect \citeauthoryear {%
Dayan%
}{%
Dayan%
}{%
{\protect \APACyear {1993}}%
}]{%
Daya93}
\APACinsertmetastar {%
Daya93}%
\begin{APACrefauthors}%
Dayan, P.%
\end{APACrefauthors}%
\unskip\
\newblock
\APACrefYearMonthDay{1993}{}{}.
\newblock
{\BBOQ}\APACrefatitle {Improving generalization for temporal difference
  learning: The successor representation} {Improving generalization for
  temporal difference learning: The successor representation}.{\BBCQ}
\newblock
\APACjournalVolNumPages{Neural Computation}{5}{4}{613--624}.
\PrintBackRefs{\CurrentBib}

\bibitem [\protect \citeauthoryear {%
Gershman%
}{%
Gershman%
}{%
{\protect \APACyear {2017}}%
}]{%
Gers17}
\APACinsertmetastar {%
Gers17}%
\begin{APACrefauthors}%
Gershman, S\BPBI J.%
\end{APACrefauthors}%
\unskip\
\newblock
\APACrefYearMonthDay{2017}{}{}.
\newblock
{\BBOQ}\APACrefatitle {Predicting the past, remembering the future} {Predicting
  the past, remembering the future}.{\BBCQ}
\newblock
\APACjournalVolNumPages{Current opinion in behavioral sciences}{17}{}{7--13}.
\PrintBackRefs{\CurrentBib}

\bibitem [\protect \citeauthoryear {%
Gershman%
\ \BBA {} Daw%
}{%
Gershman%
\ \BBA {} Daw%
}{%
{\protect \APACyear {2017}}%
}]{%
GersDaw17}
\APACinsertmetastar {%
GersDaw17}%
\begin{APACrefauthors}%
Gershman, S\BPBI J.%
\BCBT {}\ \BBA {} Daw, N\BPBI D.%
\end{APACrefauthors}%
\unskip\
\newblock
\APACrefYearMonthDay{2017}{}{}.
\newblock
{\BBOQ}\APACrefatitle {Reinforcement learning and episodic memory in humans and
  animals: An integrative framework} {Reinforcement learning and episodic
  memory in humans and animals: An integrative framework}.{\BBCQ}
\newblock
\APACjournalVolNumPages{Annual Review of Psychology}{68}{}{101--128}.
\PrintBackRefs{\CurrentBib}

\bibitem [\protect \citeauthoryear {%
Gershman%
, Moore%
, Todd%
, Norman%
\BCBL {}\ \BBA {} Sederberg%
}{%
Gershman%
\ \protect \BOthers {.}}{%
{\protect \APACyear {2012}}%
}]{%
GersEtal12}
\APACinsertmetastar {%
GersEtal12}%
\begin{APACrefauthors}%
Gershman, S\BPBI J.%
, Moore, C\BPBI D.%
, Todd, M\BPBI T.%
, Norman, K\BPBI A.%
\BCBL {}\ \BBA {} Sederberg, P\BPBI B.%
\end{APACrefauthors}%
\unskip\
\newblock
\APACrefYearMonthDay{2012}{}{}.
\newblock
{\BBOQ}\APACrefatitle {The successor representation and temporal context} {The
  successor representation and temporal context}.{\BBCQ}
\newblock
\APACjournalVolNumPages{Neural Computation}{24}{6}{1553--1568}.
\PrintBackRefs{\CurrentBib}

\bibitem [\protect \citeauthoryear {%
Green%
\ \BBA {} Myerson%
}{%
Green%
\ \BBA {} Myerson%
}{%
{\protect \APACyear {1996}}%
}]{%
GreeMyer96}
\APACinsertmetastar {%
GreeMyer96}%
\begin{APACrefauthors}%
Green, L.%
\BCBT {}\ \BBA {} Myerson, J.%
\end{APACrefauthors}%
\unskip\
\newblock
\APACrefYearMonthDay{1996}{}{}.
\newblock
{\BBOQ}\APACrefatitle {Exponential versus hyperbolic discounting of delayed
  outcomes: Risk and waiting time} {Exponential versus hyperbolic discounting
  of delayed outcomes: Risk and waiting time}.{\BBCQ}
\newblock
\APACjournalVolNumPages{American Zoologist}{36}{4}{496--505}.
\PrintBackRefs{\CurrentBib}

\bibitem [\protect \citeauthoryear {%
Green%
\ \BBA {} Myerson%
}{%
Green%
\ \BBA {} Myerson%
}{%
{\protect \APACyear {2004}}%
}]{%
GreeMyer04}
\APACinsertmetastar {%
GreeMyer04}%
\begin{APACrefauthors}%
Green, L.%
\BCBT {}\ \BBA {} Myerson, J.%
\end{APACrefauthors}%
\unskip\
\newblock
\APACrefYearMonthDay{2004}{}{}.
\newblock
{\BBOQ}\APACrefatitle {A discounting framework for choice with delayed and
  probabilistic rewards.} {A discounting framework for choice with delayed and
  probabilistic rewards.}{\BBCQ}
\newblock
\APACjournalVolNumPages{Psychological bulletin}{130}{5}{769}.
\PrintBackRefs{\CurrentBib}

\bibitem [\protect \citeauthoryear {%
Hacker%
}{%
Hacker%
}{%
{\protect \APACyear {1980}}%
}]{%
Hack80}
\APACinsertmetastar {%
Hack80}%
\begin{APACrefauthors}%
Hacker, M\BPBI J.%
\end{APACrefauthors}%
\unskip\
\newblock
\APACrefYearMonthDay{1980}{}{}.
\newblock
{\BBOQ}\APACrefatitle {Speed and accuracy of recency judgments for events in
  short-term memory.} {Speed and accuracy of recency judgments for events in
  short-term memory.}{\BBCQ}
\newblock
\APACjournalVolNumPages{Journal of Experimental Psychology: Human Learning and
  Memory}{15}{}{846-858}.
\PrintBackRefs{\CurrentBib}

\bibitem [\protect \citeauthoryear {%
Hassabis%
, Kumaran%
, Vann%
\BCBL {}\ \BBA {} Maguire%
}{%
Hassabis%
\ \protect \BOthers {.}}{%
{\protect \APACyear {2007}}%
}]{%
HassEtal07a}
\APACinsertmetastar {%
HassEtal07a}%
\begin{APACrefauthors}%
Hassabis, D.%
, Kumaran, D.%
, Vann, S\BPBI D.%
\BCBL {}\ \BBA {} Maguire, E\BPBI A.%
\end{APACrefauthors}%
\unskip\
\newblock
\APACrefYearMonthDay{2007}{}{}.
\newblock
{\BBOQ}\APACrefatitle {Patients with hippocampal amnesia cannot imagine new
  experiences} {Patients with hippocampal amnesia cannot imagine new
  experiences}.{\BBCQ}
\newblock
\APACjournalVolNumPages{Proceedings of the National Academy of Sciences
  USA}{104}{5}{1726-31}.
\newblock
\begin{APACrefDOI} \doi{10.1073/pnas.0610561104} \end{APACrefDOI}
\PrintBackRefs{\CurrentBib}

\bibitem [\protect \citeauthoryear {%
Hayden%
}{%
Hayden%
}{%
{\protect \APACyear {2016}}%
}]{%
Hayd16}
\APACinsertmetastar {%
Hayd16}%
\begin{APACrefauthors}%
Hayden, B\BPBI Y.%
\end{APACrefauthors}%
\unskip\
\newblock
\APACrefYearMonthDay{2016}{}{}.
\newblock
{\BBOQ}\APACrefatitle {Time discounting and time preference in animals: a
  critical review} {Time discounting and time preference in animals: a critical
  review}.{\BBCQ}
\newblock
\APACjournalVolNumPages{Psychonomic bulletin \& review}{23}{1}{39--53}.
\PrintBackRefs{\CurrentBib}

\bibitem [\protect \citeauthoryear {%
Hockley%
}{%
Hockley%
}{%
{\protect \APACyear {1984}}%
}]{%
Hock84}
\APACinsertmetastar {%
Hock84}%
\begin{APACrefauthors}%
Hockley, W\BPBI E.%
\end{APACrefauthors}%
\unskip\
\newblock
\APACrefYearMonthDay{1984}{}{}.
\newblock
{\BBOQ}\APACrefatitle {Analysis of response time distributions in the study of
  cognitive processes} {Analysis of response time distributions in the study of
  cognitive processes}.{\BBCQ}
\newblock
\APACjournalVolNumPages{Journal of Experimental Psychology: Learning, Memory,
  and Cognition}{10}{4}{598-615}.
\PrintBackRefs{\CurrentBib}

\bibitem [\protect \citeauthoryear {%
Howard%
, Fotedar%
, Datey%
\BCBL {}\ \BBA {} Hasselmo%
}{%
Howard%
\ \protect \BOthers {.}}{%
{\protect \APACyear {2005}}%
}]{%
HowaEtal05}
\APACinsertmetastar {%
HowaEtal05}%
\begin{APACrefauthors}%
Howard, M\BPBI W.%
, Fotedar, M\BPBI S.%
, Datey, A\BPBI V.%
\BCBL {}\ \BBA {} Hasselmo, M\BPBI E.%
\end{APACrefauthors}%
\unskip\
\newblock
\APACrefYearMonthDay{2005}{}{}.
\newblock
{\BBOQ}\APACrefatitle {The temporal context model in spatial navigation and
  relational learning: Toward a common explanation of medial temporal lobe
  function across domains} {The temporal context model in spatial navigation
  and relational learning: Toward a common explanation of medial temporal lobe
  function across domains}.{\BBCQ}
\newblock
\APACjournalVolNumPages{Psychological Review}{112}{1}{75-116}.
\PrintBackRefs{\CurrentBib}

\bibitem [\protect \citeauthoryear {%
Howard%
\ \BBA {} Kahana%
}{%
Howard%
\ \BBA {} Kahana%
}{%
{\protect \APACyear {2002}}%
}]{%
HowaKaha02a}
\APACinsertmetastar {%
HowaKaha02a}%
\begin{APACrefauthors}%
Howard, M\BPBI W.%
\BCBT {}\ \BBA {} Kahana, M\BPBI J.%
\end{APACrefauthors}%
\unskip\
\newblock
\APACrefYearMonthDay{2002}{}{}.
\newblock
{\BBOQ}\APACrefatitle {A Distributed Representation of Temporal Context} {A
  distributed representation of temporal context}.{\BBCQ}
\newblock
\APACjournalVolNumPages{Journal of Mathematical Psychology}{46}{3}{269-299}.
\PrintBackRefs{\CurrentBib}

\bibitem [\protect \citeauthoryear {%
Howard%
\ \protect \BOthers {.}}{%
Howard%
\ \protect \BOthers {.}}{%
{\protect \APACyear {2014}}%
}]{%
HowaEtal14}
\APACinsertmetastar {%
HowaEtal14}%
\begin{APACrefauthors}%
Howard, M\BPBI W.%
, Mac{D}onald, C\BPBI J.%
, Tiganj, Z.%
, Shankar, K\BPBI H.%
, Du, Q.%
, Hasselmo, M\BPBI E.%
\BCBL {}\ \BBA {} Eichenbaum, H.%
\end{APACrefauthors}%
\unskip\
\newblock
\APACrefYearMonthDay{2014}{}{}.
\newblock
{\BBOQ}\APACrefatitle {A unified mathematical framework for coding time, space,
  and sequences in the hippocampal region} {A unified mathematical framework
  for coding time, space, and sequences in the hippocampal region}.{\BBCQ}
\newblock
\APACjournalVolNumPages{Journal of Neuroscience}{34}{13}{4692-707}.
\newblock
\begin{APACrefDOI} \doi{10.1523/JNEUROSCI.5808-12.2014} \end{APACrefDOI}
\PrintBackRefs{\CurrentBib}

\bibitem [\protect \citeauthoryear {%
Howard%
\ \BBA {} Shankar%
}{%
Howard%
\ \BBA {} Shankar%
}{%
{\protect \APACyear {2018}}%
}]{%
HowaShan18}
\APACinsertmetastar {%
HowaShan18}%
\begin{APACrefauthors}%
Howard, M\BPBI W.%
\BCBT {}\ \BBA {} Shankar, K\BPBI H.%
\end{APACrefauthors}%
\unskip\
\newblock
\APACrefYearMonthDay{2018}{}{}.
\newblock
{\BBOQ}\APACrefatitle {Neural Scaling Laws for an Uncertain World} {Neural
  scaling laws for an uncertain world}.{\BBCQ}
\newblock
\APACjournalVolNumPages{Psychologial Review}{125}{}{47-58}.
\newblock
\begin{APACrefDOI} \doi{10.1037/rev0000081} \end{APACrefDOI}
\PrintBackRefs{\CurrentBib}

\bibitem [\protect \citeauthoryear {%
Howard%
, Shankar%
, Aue%
\BCBL {}\ \BBA {} Criss%
}{%
Howard%
\ \protect \BOthers {.}}{%
{\protect \APACyear {2015}}%
}]{%
HowaEtal15}
\APACinsertmetastar {%
HowaEtal15}%
\begin{APACrefauthors}%
Howard, M\BPBI W.%
, Shankar, K\BPBI H.%
, Aue, W.%
\BCBL {}\ \BBA {} Criss, A\BPBI H.%
\end{APACrefauthors}%
\unskip\
\newblock
\APACrefYearMonthDay{2015}{}{}.
\newblock
{\BBOQ}\APACrefatitle {A distributed representation of internal time} {A
  distributed representation of internal time}.{\BBCQ}
\newblock
\APACjournalVolNumPages{Psychological Review}{122}{1}{24-53}.
\PrintBackRefs{\CurrentBib}

\bibitem [\protect \citeauthoryear {%
Jin%
, Fujii%
\BCBL {}\ \BBA {} Graybiel%
}{%
Jin%
\ \protect \BOthers {.}}{%
{\protect \APACyear {2009}}%
}]{%
JinEtal09}
\APACinsertmetastar {%
JinEtal09}%
\begin{APACrefauthors}%
Jin, D\BPBI Z.%
, Fujii, N.%
\BCBL {}\ \BBA {} Graybiel, A\BPBI M.%
\end{APACrefauthors}%
\unskip\
\newblock
\APACrefYearMonthDay{2009}{}{}.
\newblock
{\BBOQ}\APACrefatitle {Neural representation of time in cortico-basal ganglia
  circuits} {Neural representation of time in cortico-basal ganglia
  circuits}.{\BBCQ}
\newblock
\APACjournalVolNumPages{Proceedings of the National Academy of
  Sciences}{106}{45}{19156--19161}.
\PrintBackRefs{\CurrentBib}

\bibitem [\protect \citeauthoryear {%
Kurth-Nelson%
\ \BBA {} Redish%
}{%
Kurth-Nelson%
\ \BBA {} Redish%
}{%
{\protect \APACyear {2009}}%
}]{%
KurtEtal09}
\APACinsertmetastar {%
KurtEtal09}%
\begin{APACrefauthors}%
Kurth-Nelson, Z.%
\BCBT {}\ \BBA {} Redish, A\BPBI D.%
\end{APACrefauthors}%
\unskip\
\newblock
\APACrefYearMonthDay{2009}{}{}.
\newblock
{\BBOQ}\APACrefatitle {Temporal-difference reinforcement learning with
  distributed representations} {Temporal-difference reinforcement learning with
  distributed representations}.{\BBCQ}
\newblock
\APACjournalVolNumPages{PLoS One}{4}{10}{e7362}.
\PrintBackRefs{\CurrentBib}

\bibitem [\protect \citeauthoryear {%
Lengyel%
\ \BBA {} Dayan%
}{%
Lengyel%
\ \BBA {} Dayan%
}{%
{\protect \APACyear {2008}}%
}]{%
LengDaya07}
\APACinsertmetastar {%
LengDaya07}%
\begin{APACrefauthors}%
Lengyel, M.%
\BCBT {}\ \BBA {} Dayan, P.%
\end{APACrefauthors}%
\unskip\
\newblock
\APACrefYearMonthDay{2008}{}{}.
\newblock
{\BBOQ}\APACrefatitle {Hippocampal contributions to control: the third way}
  {Hippocampal contributions to control: the third way}.{\BBCQ}
\newblock
\BIn{} \APACrefbtitle {Advances in neural information processing systems}
  {Advances in neural information processing systems}\ (\BPGS\ 889--896).
\PrintBackRefs{\CurrentBib}

\bibitem [\protect \citeauthoryear {%
Lisman%
, Schulman%
\BCBL {}\ \BBA {} Cline%
}{%
Lisman%
\ \protect \BOthers {.}}{%
{\protect \APACyear {2002}}%
}]{%
LismEtal02}
\APACinsertmetastar {%
LismEtal02}%
\begin{APACrefauthors}%
Lisman, J.%
, Schulman, H.%
\BCBL {}\ \BBA {} Cline, H.%
\end{APACrefauthors}%
\unskip\
\newblock
\APACrefYearMonthDay{2002}{}{}.
\newblock
{\BBOQ}\APACrefatitle {The molecular basis of CaMKII function in synaptic and
  behavioural memory} {The molecular basis of camkii function in synaptic and
  behavioural memory}.{\BBCQ}
\newblock
\APACjournalVolNumPages{Nature Reviews Neuroscience}{3}{3}{175--190}.
\PrintBackRefs{\CurrentBib}

\bibitem [\protect \citeauthoryear {%
Liu%
, Tiganj%
, Hasselmo%
\BCBL {}\ \BBA {} Howard%
}{%
Liu%
\ \protect \BOthers {.}}{%
{\protect \APACyear {{\protect \BIP {}}}}%
}]{%
LiuEtal18}
\APACinsertmetastar {%
LiuEtal18}%
\begin{APACrefauthors}%
Liu, Y.%
, Tiganj, Z.%
, Hasselmo, M\BPBI E.%
\BCBL {}\ \BBA {} Howard, M\BPBI W.%
\end{APACrefauthors}%
\unskip\
\newblock
\APACrefYearMonthDay{{\protect \BIP {}}}{}{}.
\newblock
{\BBOQ}\APACrefatitle {Biological Simulation of Scale-invariant Time Cells
  Biological Simulation of Scale-invariant Time Cells} {Biological simulation
  of scale-invariant time cells biological simulation of scale-invariant time
  cells}.{\BBCQ}
\newblock
\APACjournalVolNumPages{Hippocampus}{}{}{}.
\PrintBackRefs{\CurrentBib}

\bibitem [\protect \citeauthoryear {%
Ludvig%
, Sutton%
\BCBL {}\ \BBA {} Kehoe%
}{%
Ludvig%
\ \protect \BOthers {.}}{%
{\protect \APACyear {2008}}%
}]{%
LudvEtal08}
\APACinsertmetastar {%
LudvEtal08}%
\begin{APACrefauthors}%
Ludvig, E\BPBI A.%
, Sutton, R\BPBI S.%
\BCBL {}\ \BBA {} Kehoe, E\BPBI J.%
\end{APACrefauthors}%
\unskip\
\newblock
\APACrefYearMonthDay{2008}{}{}.
\newblock
{\BBOQ}\APACrefatitle {Stimulus Representation and the Timing of
  Reward-Prediction Errors in Models of Dopamine System} {Stimulus
  representation and the timing of reward-prediction errors in models of
  dopamine system}.{\BBCQ}
\newblock
\APACjournalVolNumPages{Neural Computation}{20}{}{3034-3054}.
\PrintBackRefs{\CurrentBib}

\bibitem [\protect \citeauthoryear {%
Ludvig%
, Sutton%
\BCBL {}\ \BBA {} Kehoe%
}{%
Ludvig%
\ \protect \BOthers {.}}{%
{\protect \APACyear {2012}}%
}]{%
LudvEtal12}
\APACinsertmetastar {%
LudvEtal12}%
\begin{APACrefauthors}%
Ludvig, E\BPBI A.%
, Sutton, R\BPBI S.%
\BCBL {}\ \BBA {} Kehoe, E\BPBI J.%
\end{APACrefauthors}%
\unskip\
\newblock
\APACrefYearMonthDay{2012}{}{}.
\newblock
{\BBOQ}\APACrefatitle {Evaluating the {TD} model of classical conditioning}
  {Evaluating the {TD} model of classical conditioning}.{\BBCQ}
\newblock
\APACjournalVolNumPages{Learning \& Behavior}{40}{3}{305--319}.
\PrintBackRefs{\CurrentBib}

\bibitem [\protect \citeauthoryear {%
MacDonald%
, Carrow%
, Place%
\BCBL {}\ \BBA {} Eichenbaum%
}{%
MacDonald%
\ \protect \BOthers {.}}{%
{\protect \APACyear {2013}}%
}]{%
MacDEtal13}
\APACinsertmetastar {%
MacDEtal13}%
\begin{APACrefauthors}%
MacDonald, C\BPBI J.%
, Carrow, S.%
, Place, R.%
\BCBL {}\ \BBA {} Eichenbaum, H.%
\end{APACrefauthors}%
\unskip\
\newblock
\APACrefYearMonthDay{2013}{}{}.
\newblock
{\BBOQ}\APACrefatitle {Distinct hippocampal time cell sequences represent odor
  memories immobilized rats.} {Distinct hippocampal time cell sequences
  represent odor memories immobilized rats.}{\BBCQ}
\newblock
\APACjournalVolNumPages{Journal of Neuroscience}{33}{36}{14607--14616}.
\PrintBackRefs{\CurrentBib}

\bibitem [\protect \citeauthoryear {%
Mac{D}onald%
, Lepage%
, Eden%
\BCBL {}\ \BBA {} Eichenbaum%
}{%
Mac{D}onald%
\ \protect \BOthers {.}}{%
{\protect \APACyear {2011}}%
}]{%
MacDEtal11}
\APACinsertmetastar {%
MacDEtal11}%
\begin{APACrefauthors}%
Mac{D}onald, C\BPBI J.%
, Lepage, K\BPBI Q.%
, Eden, U\BPBI T.%
\BCBL {}\ \BBA {} Eichenbaum, H.%
\end{APACrefauthors}%
\unskip\
\newblock
\APACrefYearMonthDay{2011}{}{}.
\newblock
{\BBOQ}\APACrefatitle {Hippocampal ``Time Cells'' Bridge the Gap in Memory for
  Discontiguous Events} {Hippocampal ``time cells'' bridge the gap in memory
  for discontiguous events}.{\BBCQ}
\newblock
\APACjournalVolNumPages{Neuron}{71}{4}{737-749}.
\PrintBackRefs{\CurrentBib}

\bibitem [\protect \citeauthoryear {%
Mau%
\ \protect \BOthers {.}}{%
Mau%
\ \protect \BOthers {.}}{%
{\protect \APACyear {2018}}%
}]{%
MauEtal18}
\APACinsertmetastar {%
MauEtal18}%
\begin{APACrefauthors}%
Mau, W.%
, Sullivan, D\BPBI W.%
, Kinsky, N\BPBI R.%
, Hasselmo, M\BPBI E.%
, Howard, M\BPBI W.%
\BCBL {}\ \BBA {} Eichenbaum, H.%
\end{APACrefauthors}%
\unskip\
\newblock
\APACrefYearMonthDay{2018}{}{}.
\newblock
{\BBOQ}\APACrefatitle {The same hippocampal CA1 population simultaneously codes
  temporal information over multiple timescales} {The same hippocampal ca1
  population simultaneously codes temporal information over multiple
  timescales}.{\BBCQ}
\newblock
\APACjournalVolNumPages{Current Biology}{28}{10}{1499--1508}.
\PrintBackRefs{\CurrentBib}

\bibitem [\protect \citeauthoryear {%
McElree%
\ \BBA {} Dosher%
}{%
McElree%
\ \BBA {} Dosher%
}{%
{\protect \APACyear {1993}}%
}]{%
McElDosh93}
\APACinsertmetastar {%
McElDosh93}%
\begin{APACrefauthors}%
McElree, B.%
\BCBT {}\ \BBA {} Dosher, B\BPBI A.%
\end{APACrefauthors}%
\unskip\
\newblock
\APACrefYearMonthDay{1993}{}{}.
\newblock
{\BBOQ}\APACrefatitle {Serial recovery processes in the recovery of order
  information.} {Serial recovery processes in the recovery of order
  information.}{\BBCQ}
\newblock
\APACjournalVolNumPages{Journal of Experimental Psychology:
  General}{122}{}{291-315}.
\PrintBackRefs{\CurrentBib}

\bibitem [\protect \citeauthoryear {%
Mello%
, Soares%
\BCBL {}\ \BBA {} Paton%
}{%
Mello%
\ \protect \BOthers {.}}{%
{\protect \APACyear {2015}}%
}]{%
MellEtal15}
\APACinsertmetastar {%
MellEtal15}%
\begin{APACrefauthors}%
Mello, G\BPBI B\BPBI M.%
, Soares, S.%
\BCBL {}\ \BBA {} Paton, J\BPBI J.%
\end{APACrefauthors}%
\unskip\
\newblock
\APACrefYearMonthDay{2015}{}{}.
\newblock
{\BBOQ}\APACrefatitle {{A Scalable Population Code for Time in the Striatum}}
  {{A Scalable Population Code for Time in the Striatum}}.{\BBCQ}
\newblock
\APACjournalVolNumPages{Current Biology}{25}{9}{1113--1122}.
\PrintBackRefs{\CurrentBib}

\bibitem [\protect \citeauthoryear {%
Mnih%
\ \protect \BOthers {.}}{%
Mnih%
\ \protect \BOthers {.}}{%
{\protect \APACyear {2015}}%
}]{%
MnihEtal15}
\APACinsertmetastar {%
MnihEtal15}%
\begin{APACrefauthors}%
Mnih, V.%
, Kavukcuoglu, K.%
, Silver, D.%
, Rusu, A\BPBI A.%
, Veness, J.%
, Bellemare, M\BPBI G.%
\BDBL {}others%
\end{APACrefauthors}%
\unskip\
\newblock
\APACrefYearMonthDay{2015}{}{}.
\newblock
{\BBOQ}\APACrefatitle {Human-level control through deep reinforcement learning}
  {Human-level control through deep reinforcement learning}.{\BBCQ}
\newblock
\APACjournalVolNumPages{Nature}{518}{7540}{529--533}.
\PrintBackRefs{\CurrentBib}

\bibitem [\protect \citeauthoryear {%
Momennejad%
\ \protect \BOthers {.}}{%
Momennejad%
\ \protect \BOthers {.}}{%
{\protect \APACyear {2017}}%
}]{%
MomeEtal17}
\APACinsertmetastar {%
MomeEtal17}%
\begin{APACrefauthors}%
Momennejad, I.%
, Russek, E\BPBI M.%
, Cheong, J\BPBI H.%
, Botvinick, M\BPBI M.%
, Daw, N.%
\BCBL {}\ \BBA {} Gershman, S\BPBI J.%
\end{APACrefauthors}%
\unskip\
\newblock
\APACrefYearMonthDay{2017}{}{}.
\newblock
{\BBOQ}\APACrefatitle {The successor representation in human reinforcement
  learning} {The successor representation in human reinforcement
  learning}.{\BBCQ}
\newblock
\APACjournalVolNumPages{Nature Human Behaviour}{1}{9}{680}.
\PrintBackRefs{\CurrentBib}

\bibitem [\protect \citeauthoryear {%
Muter%
}{%
Muter%
}{%
{\protect \APACyear {1979}}%
}]{%
Mute79}
\APACinsertmetastar {%
Mute79}%
\begin{APACrefauthors}%
Muter, P.%
\end{APACrefauthors}%
\unskip\
\newblock
\APACrefYearMonthDay{1979}{}{}.
\newblock
{\BBOQ}\APACrefatitle {Response latencies in discriminations of recency.}
  {Response latencies in discriminations of recency.}{\BBCQ}
\newblock
\APACjournalVolNumPages{Journal of Experimental Psychology: Human Learning and
  Memory}{5}{}{160-169}.
\PrintBackRefs{\CurrentBib}

\bibitem [\protect \citeauthoryear {%
Palombo%
, Keane%
\BCBL {}\ \BBA {} Verfaellie%
}{%
Palombo%
\ \protect \BOthers {.}}{%
{\protect \APACyear {2015}}%
}]{%
PaloEtal15}
\APACinsertmetastar {%
PaloEtal15}%
\begin{APACrefauthors}%
Palombo, D\BPBI J.%
, Keane, M\BPBI M.%
\BCBL {}\ \BBA {} Verfaellie, M.%
\end{APACrefauthors}%
\unskip\
\newblock
\APACrefYearMonthDay{2015}{}{}.
\newblock
{\BBOQ}\APACrefatitle {The medial temporal lobes are critical for reward-based
  decision making under conditions that promote episodic future thinking} {The
  medial temporal lobes are critical for reward-based decision making under
  conditions that promote episodic future thinking}.{\BBCQ}
\newblock
\APACjournalVolNumPages{Hippocampus}{25}{3}{345--353}.
\PrintBackRefs{\CurrentBib}

\bibitem [\protect \citeauthoryear {%
Pastalkova%
, Itskov%
, Amarasingham%
\BCBL {}\ \BBA {} Buzsaki%
}{%
Pastalkova%
\ \protect \BOthers {.}}{%
{\protect \APACyear {2008}}%
}]{%
PastEtal08}
\APACinsertmetastar {%
PastEtal08}%
\begin{APACrefauthors}%
Pastalkova, E.%
, Itskov, V.%
, Amarasingham, A.%
\BCBL {}\ \BBA {} Buzsaki, G.%
\end{APACrefauthors}%
\unskip\
\newblock
\APACrefYearMonthDay{2008}{}{}.
\newblock
{\BBOQ}\APACrefatitle {Internally generated cell assembly sequences in the rat
  hippocampus.} {Internally generated cell assembly sequences in the rat
  hippocampus.}{\BBCQ}
\newblock
\APACjournalVolNumPages{Science}{321}{5894}{1322-7}.
\PrintBackRefs{\CurrentBib}

\bibitem [\protect \citeauthoryear {%
Polyn%
, Norman%
\BCBL {}\ \BBA {} Kahana%
}{%
Polyn%
\ \protect \BOthers {.}}{%
{\protect \APACyear {2009}}%
}]{%
PolyEtal09}
\APACinsertmetastar {%
PolyEtal09}%
\begin{APACrefauthors}%
Polyn, S\BPBI M.%
, Norman, K\BPBI A.%
\BCBL {}\ \BBA {} Kahana, M\BPBI J.%
\end{APACrefauthors}%
\unskip\
\newblock
\APACrefYearMonthDay{2009}{}{}.
\newblock
{\BBOQ}\APACrefatitle {A context maintenance and retrieval model of
  organizational processes in free recall} {A context maintenance and retrieval
  model of organizational processes in free recall}.{\BBCQ}
\newblock
\APACjournalVolNumPages{Psychological Review}{116}{}{129-156}.
\PrintBackRefs{\CurrentBib}

\bibitem [\protect \citeauthoryear {%
Post%
}{%
Post%
}{%
{\protect \APACyear {1930}}%
}]{%
Post30}
\APACinsertmetastar {%
Post30}%
\begin{APACrefauthors}%
Post, E.%
\end{APACrefauthors}%
\unskip\
\newblock
\APACrefYearMonthDay{1930}{}{}.
\newblock
{\BBOQ}\APACrefatitle {Generalized Differentiation} {Generalized
  differentiation}.{\BBCQ}
\newblock
\APACjournalVolNumPages{Transactions of the American Mathematical
  Society}{32}{}{723-781}.
\PrintBackRefs{\CurrentBib}

\bibitem [\protect \citeauthoryear {%
Salz%
\ \protect \BOthers {.}}{%
Salz%
\ \protect \BOthers {.}}{%
{\protect \APACyear {2016}}%
}]{%
SalzEtal16}
\APACinsertmetastar {%
SalzEtal16}%
\begin{APACrefauthors}%
Salz, D\BPBI M.%
, Tiganj, Z.%
, Khasnabish, S.%
, Kohley, A.%
, Sheehan, D.%
, Howard, M\BPBI W.%
\BCBL {}\ \BBA {} Eichenbaum, H.%
\end{APACrefauthors}%
\unskip\
\newblock
\APACrefYearMonthDay{2016}{}{}.
\newblock
{\BBOQ}\APACrefatitle {Time Cells in Hippocampal Area CA3} {Time cells in
  hippocampal area ca3}.{\BBCQ}
\newblock
\APACjournalVolNumPages{The Journal of Neuroscience}{36}{28}{7476--7484}.
\PrintBackRefs{\CurrentBib}

\bibitem [\protect \citeauthoryear {%
Schacter%
, Addis%
\BCBL {}\ \BBA {} Buckner%
}{%
Schacter%
\ \protect \BOthers {.}}{%
{\protect \APACyear {2007}}%
}]{%
SchaEtal07}
\APACinsertmetastar {%
SchaEtal07}%
\begin{APACrefauthors}%
Schacter, D\BPBI L.%
, Addis, D\BPBI R.%
\BCBL {}\ \BBA {} Buckner, R\BPBI L.%
\end{APACrefauthors}%
\unskip\
\newblock
\APACrefYearMonthDay{2007}{}{}.
\newblock
{\BBOQ}\APACrefatitle {Remembering the past to imagine the future: the
  prospective brain} {Remembering the past to imagine the future: the
  prospective brain}.{\BBCQ}
\newblock
\APACjournalVolNumPages{Nature Reviews, Neuroscience}{8}{9}{657-661}.
\PrintBackRefs{\CurrentBib}

\bibitem [\protect \citeauthoryear {%
Schultz%
, Dayan%
\BCBL {}\ \BBA {} Montague%
}{%
Schultz%
\ \protect \BOthers {.}}{%
{\protect \APACyear {1997}}%
}]{%
SchuEtal97}
\APACinsertmetastar {%
SchuEtal97}%
\begin{APACrefauthors}%
Schultz, W.%
, Dayan, P.%
\BCBL {}\ \BBA {} Montague, P\BPBI R.%
\end{APACrefauthors}%
\unskip\
\newblock
\APACrefYearMonthDay{1997}{}{}.
\newblock
{\BBOQ}\APACrefatitle {A neural substrate of prediction and reward} {A neural
  substrate of prediction and reward}.{\BBCQ}
\newblock
\APACjournalVolNumPages{Science}{275}{}{1593-1599}.
\PrintBackRefs{\CurrentBib}

\bibitem [\protect \citeauthoryear {%
Sederberg%
, Howard%
\BCBL {}\ \BBA {} Kahana%
}{%
Sederberg%
\ \protect \BOthers {.}}{%
{\protect \APACyear {2008}}%
}]{%
SedeEtal08}
\APACinsertmetastar {%
SedeEtal08}%
\begin{APACrefauthors}%
Sederberg, P\BPBI B.%
, Howard, M\BPBI W.%
\BCBL {}\ \BBA {} Kahana, M\BPBI J.%
\end{APACrefauthors}%
\unskip\
\newblock
\APACrefYearMonthDay{2008}{}{}.
\newblock
{\BBOQ}\APACrefatitle {A context-based theory of recency and contiguity in free
  recall} {A context-based theory of recency and contiguity in free
  recall}.{\BBCQ}
\newblock
\APACjournalVolNumPages{Psychological Review}{115}{}{893-912}.
\PrintBackRefs{\CurrentBib}

\bibitem [\protect \citeauthoryear {%
Shankar%
\ \BBA {} Howard%
}{%
Shankar%
\ \BBA {} Howard%
}{%
{\protect \APACyear {2012}}%
}]{%
ShanHowa12}
\APACinsertmetastar {%
ShanHowa12}%
\begin{APACrefauthors}%
Shankar, K\BPBI H.%
\BCBT {}\ \BBA {} Howard, M\BPBI W.%
\end{APACrefauthors}%
\unskip\
\newblock
\APACrefYearMonthDay{2012}{}{}.
\newblock
{\BBOQ}\APACrefatitle {A scale-invariant internal representation of time} {A
  scale-invariant internal representation of time}.{\BBCQ}
\newblock
\APACjournalVolNumPages{Neural Computation}{24}{1}{134-193}.
\PrintBackRefs{\CurrentBib}

\bibitem [\protect \citeauthoryear {%
Shankar%
\ \BBA {} Howard%
}{%
Shankar%
\ \BBA {} Howard%
}{%
{\protect \APACyear {2013}}%
}]{%
ShanHowa13}
\APACinsertmetastar {%
ShanHowa13}%
\begin{APACrefauthors}%
Shankar, K\BPBI H.%
\BCBT {}\ \BBA {} Howard, M\BPBI W.%
\end{APACrefauthors}%
\unskip\
\newblock
\APACrefYearMonthDay{2013}{}{}.
\newblock
{\BBOQ}\APACrefatitle {Optimally fuzzy temporal memory} {Optimally fuzzy
  temporal memory}.{\BBCQ}
\newblock
\APACjournalVolNumPages{Journal of Machine Learning Research}{14}{}{3753-3780}.
\PrintBackRefs{\CurrentBib}

\bibitem [\protect \citeauthoryear {%
Singh%
\ \BBA {} Howard%
}{%
Singh%
\ \BBA {} Howard%
}{%
{\protect \APACyear {2017}}%
}]{%
SingHowa17a}
\APACinsertmetastar {%
SingHowa17a}%
\begin{APACrefauthors}%
Singh, I.%
\BCBT {}\ \BBA {} Howard, M\BPBI W.%
\end{APACrefauthors}%
\unskip\
\newblock
\APACrefYearMonthDay{2017}{}{}.
\newblock
{\BBOQ}\APACrefatitle {Scanning along a compressed timeline of the future}
  {Scanning along a compressed timeline of the future}.{\BBCQ}
\newblock
\APACjournalVolNumPages{bioRxiv}{}{}{229617}.
\PrintBackRefs{\CurrentBib}

\bibitem [\protect \citeauthoryear {%
Stachenfeld%
, Botvinick%
\BCBL {}\ \BBA {} Gershman%
}{%
Stachenfeld%
\ \protect \BOthers {.}}{%
{\protect \APACyear {2016}}%
}]{%
StacEtal16}
\APACinsertmetastar {%
StacEtal16}%
\begin{APACrefauthors}%
Stachenfeld, K\BPBI L.%
, Botvinick, M\BPBI M.%
\BCBL {}\ \BBA {} Gershman, S\BPBI J.%
\end{APACrefauthors}%
\unskip\
\newblock
\APACrefYearMonthDay{2016}{}{}.
\newblock
{\BBOQ}\APACrefatitle {The hippocampus as a predictive map} {The hippocampus as
  a predictive map}.{\BBCQ}
\newblock
\APACjournalVolNumPages{bioRxiv}{}{}{097170}.
\PrintBackRefs{\CurrentBib}

\bibitem [\protect \citeauthoryear {%
Sutton%
}{%
Sutton%
}{%
{\protect \APACyear {1995}}%
}]{%
Sutt95}
\APACinsertmetastar {%
Sutt95}%
\begin{APACrefauthors}%
Sutton, R\BPBI S.%
\end{APACrefauthors}%
\unskip\
\newblock
\APACrefYearMonthDay{1995}{}{}.
\newblock
{\BBOQ}\APACrefatitle {TD models: Modeling the world at a mixture of time
  scales} {Td models: Modeling the world at a mixture of time scales}.{\BBCQ}
\newblock
\BIn{} \APACrefbtitle {ICML} {Icml}\ (\BVOL~12, \BPGS\ 531--539).
\PrintBackRefs{\CurrentBib}

\bibitem [\protect \citeauthoryear {%
Sutton%
\ \BBA {} Barto%
}{%
Sutton%
\ \BBA {} Barto%
}{%
{\protect \APACyear {1998}}%
}]{%
SuttEtal98}
\APACinsertmetastar {%
SuttEtal98}%
\begin{APACrefauthors}%
Sutton, R\BPBI S.%
\BCBT {}\ \BBA {} Barto, A\BPBI G.%
\end{APACrefauthors}%
\unskip\
\newblock
\APACrefYear{1998}.
\newblock
\APACrefbtitle {Reinforcement learning: An introduction} {Reinforcement
  learning: An introduction}\ (\BVOL~1)\ (\BNUM~1).
\newblock
\APACaddressPublisher{}{MIT press Cambridge}.
\PrintBackRefs{\CurrentBib}

\bibitem [\protect \citeauthoryear {%
Terada%
, Sakurai%
, Nakahara%
\BCBL {}\ \BBA {} Fujisawa%
}{%
Terada%
\ \protect \BOthers {.}}{%
{\protect \APACyear {2017}}%
}]{%
TeraEtal17}
\APACinsertmetastar {%
TeraEtal17}%
\begin{APACrefauthors}%
Terada, S.%
, Sakurai, Y.%
, Nakahara, H.%
\BCBL {}\ \BBA {} Fujisawa, S.%
\end{APACrefauthors}%
\unskip\
\newblock
\APACrefYearMonthDay{2017}{}{}.
\newblock
{\BBOQ}\APACrefatitle {Temporal and Rate Coding for Discrete Event Sequences in
  the Hippocampus} {Temporal and rate coding for discrete event sequences in
  the hippocampus}.{\BBCQ}
\newblock
\APACjournalVolNumPages{Neuron}{}{}{}.
\PrintBackRefs{\CurrentBib}

\bibitem [\protect \citeauthoryear {%
Tiganj%
, Cromer%
, Roy%
, Miller%
\BCBL {}\ \BBA {} Howard%
}{%
Tiganj%
\ \protect \BOthers {.}}{%
{\protect \APACyear {2018}}%
}]{%
TigaEtal18a}
\APACinsertmetastar {%
TigaEtal18a}%
\begin{APACrefauthors}%
Tiganj, Z.%
, Cromer, J\BPBI A.%
, Roy, J\BPBI E.%
, Miller, E\BPBI K.%
\BCBL {}\ \BBA {} Howard, M\BPBI W.%
\end{APACrefauthors}%
\unskip\
\newblock
\APACrefYearMonthDay{2018}{}{}.
\newblock
{\BBOQ}\APACrefatitle {Compressed Timeline of Recent Experience in Monkey
  Lateral Prefrontal Cortex} {Compressed timeline of recent experience in
  monkey lateral prefrontal cortex}.{\BBCQ}
\newblock
\APACjournalVolNumPages{Journal of cognitive neuroscience}{}{}{1--16}.
\PrintBackRefs{\CurrentBib}

\bibitem [\protect \citeauthoryear {%
Tiganj%
, Hasselmo%
\BCBL {}\ \BBA {} Howard%
}{%
Tiganj%
\ \protect \BOthers {.}}{%
{\protect \APACyear {2015}}%
}]{%
TigaEtal15}
\APACinsertmetastar {%
TigaEtal15}%
\begin{APACrefauthors}%
Tiganj, Z.%
, Hasselmo, M\BPBI E.%
\BCBL {}\ \BBA {} Howard, M\BPBI W.%
\end{APACrefauthors}%
\unskip\
\newblock
\APACrefYearMonthDay{2015}{}{}.
\newblock
{\BBOQ}\APACrefatitle {A Simple biophysically plausible model for long time
  constants in single neurons} {A simple biophysically plausible model for long
  time constants in single neurons}.{\BBCQ}
\newblock
\APACjournalVolNumPages{Hippocampus}{25}{1}{27-37}.
\PrintBackRefs{\CurrentBib}

\bibitem [\protect \citeauthoryear {%
Tiganj%
, Kim%
, Jung%
\BCBL {}\ \BBA {} Howard%
}{%
Tiganj%
\ \protect \BOthers {.}}{%
{\protect \APACyear {2016}}%
}]{%
TigaEtal16}
\APACinsertmetastar {%
TigaEtal16}%
\begin{APACrefauthors}%
Tiganj, Z.%
, Kim, J.%
, Jung, M\BPBI W.%
\BCBL {}\ \BBA {} Howard, M\BPBI W.%
\end{APACrefauthors}%
\unskip\
\newblock
\APACrefYearMonthDay{2016}{}{}.
\newblock
{\BBOQ}\APACrefatitle {Sequential firing codes for time in rodent m{PFC}}
  {Sequential firing codes for time in rodent m{PFC}}.{\BBCQ}
\newblock
\APACjournalVolNumPages{Cerebral Cortex}{}{1-9}{}.
\newblock
\begin{APACrefDOI} \doi{10.1093/cercor/bhw336} \end{APACrefDOI}
\PrintBackRefs{\CurrentBib}

\bibitem [\protect \citeauthoryear {%
Tulving%
}{%
Tulving%
}{%
{\protect \APACyear {1983}}%
}]{%
Tulv83}
\APACinsertmetastar {%
Tulv83}%
\begin{APACrefauthors}%
Tulving, E.%
\end{APACrefauthors}%
\unskip\
\newblock
\APACrefYear{1983}.
\newblock
\APACrefbtitle {Elements of Episodic Memory} {Elements of episodic memory}.
\newblock
\APACaddressPublisher{New York}{Oxford}.
\PrintBackRefs{\CurrentBib}

\bibitem [\protect \citeauthoryear {%
Tulving%
}{%
Tulving%
}{%
{\protect \APACyear {1985}}%
}]{%
Tulv85a}
\APACinsertmetastar {%
Tulv85a}%
\begin{APACrefauthors}%
Tulving, E.%
\end{APACrefauthors}%
\unskip\
\newblock
\APACrefYearMonthDay{1985}{}{}.
\newblock
{\BBOQ}\APACrefatitle {Memory and consciousness} {Memory and
  consciousness}.{\BBCQ}
\newblock
\APACjournalVolNumPages{Canadian Psychology}{26}{1}{1-12}.
\PrintBackRefs{\CurrentBib}

\bibitem [\protect \citeauthoryear {%
Wimmer%
\ \BBA {} Shohamy%
}{%
Wimmer%
\ \BBA {} Shohamy%
}{%
{\protect \APACyear {2012}}%
}]{%
WimmShoh12}
\APACinsertmetastar {%
WimmShoh12}%
\begin{APACrefauthors}%
Wimmer, G\BPBI E.%
\BCBT {}\ \BBA {} Shohamy, D.%
\end{APACrefauthors}%
\unskip\
\newblock
\APACrefYearMonthDay{2012}{}{}.
\newblock
{\BBOQ}\APACrefatitle {Preference by association: how memory mechanisms in the
  hippocampus bias decisions} {Preference by association: how memory mechanisms
  in the hippocampus bias decisions}.{\BBCQ}
\newblock
\APACjournalVolNumPages{Science}{338}{6104}{270-3}.
\newblock
\begin{APACrefDOI} \doi{10.1126/science.1223252} \end{APACrefDOI}
\PrintBackRefs{\CurrentBib}

\end{thebibliography}

\end{document}